\pdfoutput=1

\documentclass[11pt]{article}

\usepackage[final]{acl}

\usepackage{times}
\usepackage{latexsym}
\usepackage{comment}

\usepackage[T1]{fontenc}

\usepackage[utf8]{inputenc}

\usepackage{microtype}

\usepackage{inconsolata}

\usepackage{graphicx}
\usepackage{booktabs}
\usepackage{multirow}
\usepackage{xcolor}
\usepackage{hyperref}
\usepackage{amsmath}
\usepackage{mathtools}
\usepackage{makecell}
\usepackage{ltablex}
\usepackage{longtable}
\usepackage{supertabular,booktabs}
\usepackage{xltabular}
\usepackage{tikz}
\usepackage{pgfplots}
\usetikzlibrary{patterns}
\usepackage{xspace}
\usepackage{booktabs,siunitx}
\usepackage{pgf-pie}

\title{Faithful, Unfaithful or Ambiguous? Multi-Agent Debate with Initial Stance for Summary Evaluation}

\author{
 \textbf{Mahnaz Koupaee\textsuperscript{1}\thanks{Work done as an intern at Amazon.}},
 \textbf{Jake W. Vincent\textsuperscript{2}},
 \textbf{Saab Mansour\textsuperscript{2}},
 \textbf{Igor Shalyminov\textsuperscript{2}},
\\
 \textbf{Han He\textsuperscript{2}},
 \textbf{Hwanjun Song\textsuperscript{3}},
 \textbf{Raphael Shu\textsuperscript{2}},
 \textbf{Jianfeng He \textsuperscript{2}},
\\
 \textbf{Yi Nian\textsuperscript{2}},
 \textbf{Amy Wing-mei Wong\textsuperscript{2}},
 \textbf{Kyu J. Han\textsuperscript{2}},
 \textbf{Hang Su\textsuperscript{2}},
\\
\\
 \textsuperscript{1}Stony Brook University,
 \textsuperscript{2}Amazon,
 \textsuperscript{3}Korea Advanced Institute of Science and Technology
\\
 \texttt{
   mkoupaee@cs.stonybrook.edu
 }
}

\newcommand{\method}[0]{\textsc{Madisse}\xspace}

\begin{document}
\maketitle
\begin{abstract}
Faithfulness evaluators based on large language models (LLMs) are often fooled by the fluency of the text and struggle with identifying errors in the summaries. 
We propose an approach to summary faithfulness evaluation in which multiple LLM-based agents are assigned initial stances (regardless of what their belief might be) and forced to come up with a reason to justify the imposed belief, thus engaging in a multi-round debate to reach an agreement. The uniformly distributed initial assignments result in a greater diversity of stances leading to more meaningful debates and ultimately more errors identified.
Furthermore, by analyzing the recent faithfulness evaluation datasets, we observe that naturally, it is not always the case for a summary to be either faithful to the source document or not. We therefore introduce a new dimension, \textbf{\textit{ambiguity}}, and a detailed taxonomy to identify such special cases. Experiments demonstrate our approach can help identify ambiguities, and have even a stronger performance on non-ambiguous summaries\footnote{Code and data available at \href{https://github.com/amazon-science/madisse}{github.com/amazon-science/madisse}}.

\end{abstract}

\section{Introduction}
Summary evaluation has a long history, and over the years, different approaches have been applied to evaluate the quality of the generated summaries including n-gram based metrics \cite{lin2004rouge,papineni2002bleu}, representation-based approaches \cite{zhang2020bertscoreevaluatingtextgeneration}, finetuned specialized evaluators \cite{kryscinski2020evaluating,fabbri2022qafacteval,goyal2020evaluating,clark2023seahorse,tang2024minicheck} and human evaluation. 
With recent advancements in LLMs and their superior ability to generate fluent text, automatic summary evaluation has gained even more attention. In particular, assessing aspects like faithfulness has become more challenging due to the high fluency of LLM-generated text. 

While overlap-based metrics usually show weak correlation with human judgments \cite{liu2023g,tang2024tofueval} and finetuned approaches usually lack explainability, human evaluation is also costly with high turnaround time, low reproduciblity and low inter annotator agreement (IAA). With that said, efficient and accurate evaluation of summaries still remains a challenge. 

Automatic evaluation using LLMs have shown promising results, overcoming some of the major bottlenecks of traditional approaches in efficient evaluation of the generated summaries.
Different single-LLM and multi-LLM settings have been applied on a wide range of tasks and are shown to be strong automatic evaluators \cite{liu2023g,luo2023chatgptfactualinconsistencyevaluator,wang2023factcheck,chan2023chateval,song2024finesure}. 
But even LLMs as evaluators fail to identify a large portion of the errors and are often fooled by the fluency of the LLM-generated summaries. 
Interestingly, when told that a given summary is unfaithful, LLMs can come up with correct reasoning and arguments that they couldn't otherwise, showing their inherent potential for error detection.
To efficiently exploit the error detection capability of the LLMs to reason about the faithfulness of a given summary, we propose \textbf{\method}, a Multi-Agent Debate with Initial Stance for Summary Evaluation framework, in which LLM-based agents will be assigned opposing initial stances (either faithful or unfaithful) as their beliefs on the faithfulness quality of the summary. 
Forcing LLMs to come up with reasons to justify an initial stance might not always lead to correct prediction as the stances are random and might not be aligned with actual faithfulness labels.
Therefore, agents engage in multiple rounds of debate with each other, either support or refute others’ arguments with the aim of resolving any inconsistencies and reaching an agreement on the final label.

However, the main underlying assumption in faithfulness evaluation is that a summary \textsc{always} has a right answer and can either be classified as faithful or unfaithful which might not be the case. 
A summary can be interpreted in different correct and plausible ways and then depending on the interpretation can be seen as both faithful and unfaithful as shown in Figure \ref{fig:ambiguity-motiv}. This would lead to low IAA regardless of the quality of the evaluators as they might only think of one interpretation and base their evaluation on that. 
The possibility of a summary having multiple interpretations leading to different faithfulness evaluations can impact the conclusions regarding system performance and ranking.
We therefore introduce a new evaluation dimension, \textbf{\textit{ambiguity}}, and we define it as when a summary can have multiple correct interpretations in context of the given document leading to opposing beliefs about the faithfulness of the summary. 
An optimal faithfulness evaluator should address any ambiguities before evaluating faithfulness and the initial step in doing so is to identify such ambiguous summaries.
To facilitate this, we also provide a detailed taxonomy of ambiguities and a human annotated dataset by extending the TofuEval MeetingBank dataset \cite{tang2024tofueval} with ambiguity annotations.

Our main contributions can be summarized as follows:
(1) We propose \method, a multi-agent debate setup with initial stance for improved faithfulness evaluation leading to stronger performance compared to single-LLM and multi-LLM setups for non-ambiguous scenarios by identifying more errors;
(2) We introduce a new evaluation dimension, ambiguity, a detailed taxonomy of ambiguity types and provide ambiguity annotation on TofuEval MeetingBank dataset;
(3) We show how the debate approach can help with identifying ambiguous cases and furthermore can even have a stronger performance in terms of accuracy and increasing IAA, when evaluated on non-ambiguous summaries.

\section{Related Work}

Evaluation of summary faithfulness has been extensively studied before. We present an overview of such works, with special attention to the recent LLM-based and multi-agent approaches.

\subsection{Summary Evaluation}

Automatic n-gram based metrics such as ROUGE \cite{lin2004rouge} and BLEU \cite{papineni2002bleu} or representation-based metrics such as BERTScore \cite{zhang2020bertscoreevaluatingtextgeneration} have long been used to measure the quality of a generated summary with respect to a given reference summary (or the document). However, they have been shown to have poor correlation with human judgments \cite{gao-wan-2022-dialsummeval,Tang2023.04.22.23288967}. The reason behind that is the arrival of LLMs which have proven to be extremely good at generating text of a high quality, relevance and at the same time of enough diversity to mislead the word overlap/distance-based metrics. Moreover, the LLMs' parametric knowledge would lead to new subtleties that cannot be easily directed with the traditional automatic metrics. One of the major issues with employing LLMs as summarizers is {\it hallucination}, when the LLM generates a fact solely using its parametric knowledge and without grounding it in the source document.
Many approaches were developed to overcome those challenges in summary evaluation, which we categorize into two.
First, specialized error detectors which are trained to detect a specific type of error in the generated summary \cite{kryscinski2020evaluating, fabbri2022qafacteval, goyal2020evaluating, clark2023seahorse, tang2024minicheck}. However, these approaches require annotated data and only provide a single faithfulness label without localizing the error.
 Second, LLM-based evaluators through zero-shot prompting \cite{luo2023chatgptfactualinconsistencyevaluator, wang2023factcheck}. In these approaches, the LLMs are provided with the task description and are asked to evaluate the given text by either providing a label or a ranking. The final result can also be an aggregation of the responses from multiple LLMs that are instructed to do the same task \cite{verga2024replacing}.
 Though shown to be competitive with human evaluations, they still miss on a large portion of the errors \cite{tang2024minicheck,tang2024tofueval}.

\subsection{LLM-Based Multi-Agent Systems}

Single LLM agents have shown promising results in many tasks and applications, however, LLM-based multi-agents have been proposed to further expand their capabilities and to better leverage their expertise and skills.  
There are two main system categories: in the first category, different LLMs are asked to do the same task but either with a specific role in mind such as a critic or general public \cite{chan2023chateval} or are asked to do it using the feedback from other agents and try to modify their response with respect to other agents responses through rounds of debates \cite{du2023improving}. In this setting of peer-to-peer debaters with a judge, a known problem is the {\it degeneration of thought} when, having acquired some confidence in its stance, the debater will stick to it whether it's correct or not, making the potentially lengthy and costly further debate of little use. In this case, the diversity of the debaters' stances becomes important, and as such, \citet{DBLP:journals/corr/abs-2305-19118} assign roles (affirmative, disagreeing) to the agents in the prompts, having the judge combine all the debaters' arguments and come up with the final decision. \citet{DBLP:conf/icml/SmitGDBP24} also explore the {\it agreement modulation} technique in which they assign each debater the ratio with which it agrees with others' points of view, leading to notable performance improvements. \citet{DBLP:conf/acl/ZhangX0LHD24} explore both personality traits of the agents (easy going / overconfident) and thinking patterns (self-reflection / debating) and their contribution to the debate outcome. In the second category, multiple LLMs can collaborate together through a set of guidelines to do a task with each agent only doing a part of the job \cite{mandi2024roco, qian2024chatdev, hong2023metagpt, lan2024stance}. In this setup, a task is broken into smaller sub-tasks that require different skill set and all agents work towards reaching the broader goal by realizing their specified tasks. 
Our approach is similar to the first category in which multiple evaluators with different initial instances engage in a debate to reach a conclusion on the faithfulness of a given summary.

\section{\method}

\begin{figure*}
    \centering
    \includegraphics[width=1\linewidth]{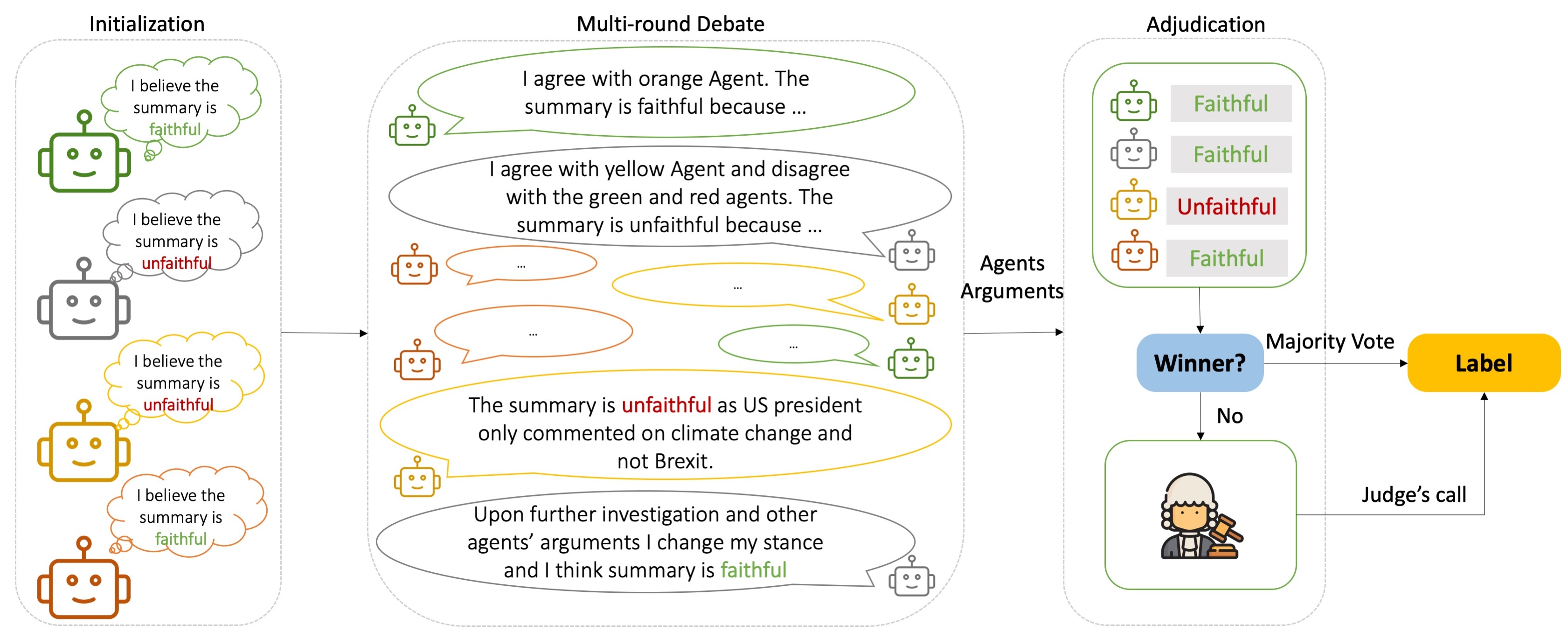}
    \vspace*{-0.7cm}
    \caption{Overview of \method, our proposed framework for automatic faithfulness evaluation. Each debate session consists of three stages: 1) \textit{stance initialization}, in which agents are assigned a belief of the summary faithfulness (faithful or unfaithful), 2) \textit{debate}, where evaluator agents engage in multiple rounds of debate to persuade each other of whether the summary is faithful or not, and 3) \textit{adjudication}, where based on the arguments from the debate, the final label is assigned to the summary. \method can have simultaneous debate sessions.}
    \label{fig:overview}
    \vspace*{-0.2cm}
\end{figure*}

\textit{Faithfulness} as a key evaluation dimension of summarization systems, measures whether the facts specified in the summary can be attributed to the source document. 
We focus on faithfulness as described above and consider summaries to be faithful if only they can be entailed from the source document\footnote{\textit{faithfulness} is different from \textit{factuality} as for factuality, it is enough for a summary to be attributed to the world knowledge~\cite{maynez-etal-2020-faithfulness}.}.
Formally, we define an evaluation model $M$ to predict whether the summary $s$ can be entailed from the source document $D$.
\[M(D, s) \in \{\text{faithful, unfaithful}\}\]
The overview of \method can be seen in Figure \ref{fig:overview}. Each \method session consists of three main stages: initialization, debate and adjudication.

In the initialization stage, a pool of evaluator agents $\mathcal{A}$ are assigned a random stance on whether they believe the summary is faithful or not. 
In the second stage, the agents engage in a debate for $n$ rounds and each agent $A_i \in \mathcal{A}$ provides arguments $U^j_i$ at each round $j$ which consists of an explanation and a label for the summary: $U^j_i = (e^j_i, l^j_i)$ where $e^j_i$ is the explanation to justify the decision and $l^j_i$ is the faithfulness label assigned to the summary at round $j$ for the $i$-th agent $A_i$. If at any round $j$, {\em all} agents agree on the final label, the debate will be stopped and the final label of the summary is determined.
If agents do not reach an agreement after $n$ rounds, the debate will stop and then the final label is determined by adjudication. Adjudicators $J_1, ..., J_k \in \mathcal{J}$ are judges responsible for checking every agent’s arguments $U_i$ and making the final call. 

In the following sections, we will detail each component of \method, describing their responsibilities and goals and how they achieve them. 

\subsection{Initialization}
A debate would be more engaging if the involved parties have conflicting overviews on the topic as they are encouraged to think deeper to come up with better arguments for their beliefs. 
This is also the case for faithfulness evaluation where arguing for conflicting opinions on faithfulness can lead to deeper understanding of the semantics of the summary and even better judgment of the faithfulness. 

One way to inject the desired diversity is to assign the evaluator agents an initial stance: $A_i \leftarrow{f_0}$. More specifically, $f_0$ will be the first argument $U^0_i$ for each agent $A_i$ which they believe is their assessment of the summary. These initial arguments will be part of the chat history for the debate stage (the initial evaluator agent prompt is shown in Table \ref{tab:debate_prompt_first_round} in Appendix \ref{app:prompts}).

We assign initial stances such that half of the evaluator agents start the debate by believing the summary is faithful and the other half believing the summary is unfaithful (uniform distribution of stances). Therefore, $U^0_i$ can be one of the two: \{\textit{The summary is faithful, The summary is unfaithful}\}. 
We later show how effective this initialization is in detecting cases that would go unnoticed otherwise.
It can also help with ambiguity detection later discussed in Section \ref{sec:ambiguity})

\subsection{Multi-Round Debate}
During each debate stage, each LLM-based evaluator agent $A_i \in \mathcal{A}$ would go over the document $D$ and the summary $s$ and look for potential inconsistencies that might be present in the summary. Each agent is also aware of the existence of other agents and they are encouraged to continue the debate with each other, specify why other agents might be right or wrong and also ask some follow-up or clarification questions. 
At each round $j$, each agent $A_i$ has access to the previous chat history and what other agents argued for. Then it generates a new argument $U^j_i$ for the current round providing a faithfulness label and why it believes the label is justified (the evaluator prompt is shown in the Appendix \ref{app:prompts} Table \ref{tab:debate_prompt}~\footnote{Note that the evaluator prompt in Table~\ref{tab:debate_prompt} is similar to the initial evaluator prompt in Table~\ref{tab:debate_prompt_first_round} except for the chat history part which is dynamic and excludes the initial imposed stances.}). This argument will be added to the chat history for the next rounds. 
Also, to remove any ordering biases, we shuffle the arguments from each round before showing the chat history to the agents for subsequent debate rounds. 

The debate stage has two main properties: guidelines and stopping criterion.
The first property borrows ideas from collaborative human workflows in which we design guidelines/rules that agents can use and refer to during the debate and making their arguments, which would help with having a more structured debate for easier reference. The stopping criterion is also required to make sure the debate will conclude and the summary is evaluated. These two properties are described in more detail in the Appendix \ref{app:guidelines} and \ref{app:stop}.
The benefits of the debate stage are two-fold. 
Not only does the debate setup provide an opportunity of collaboration among different evaluator agents towards the correct decision, it also helps with resolving inconsistencies that might occur due to stance initialization stage.

\subsection{Adjudication}

Even after rounds of debate, the evaluator agents might still disagree. However, the debate can only go on for $n$ rounds. Once the debate is over, the adjudicator module consisting of $k$ adjudicator agents $J_1, ..., J_k \in \mathcal{J}$ receives all the final arguments $U^n_i$ from the evaluator agents $\mathcal{A}$, goes over them and makes sure they are well aligned with the provided guidelines. 
Then based on the agents’ responses as well as its own judgment, the adjudicator makes the final call on the summary by providing a label as well as an explanation (the adjudicator prompt is shown in Appendix \ref{app:prompts}, Table~\ref{tab:adjudicator_prompt}). To make sure that the adjudication is not biased towards the agents’ arguments order, we use multiple adjudicators each time with a different random order and then finally do a majority voting to get the final response $U^n_k = (e^n_k, l^n_k)$ (the explanation $e^n_k$ is selected randomly from the majority vote responses).

\subsection{Simultaneous Debate Sessions}\label{app:method}
A debate among agents with adjudicators to help with final decision can result in two major type of errors; adjudicator mistake and wrong answer propagation \cite{wang2024rethinking}. The first one happens when adjudicators select the wrong option as the final response specially in cases where there are conflicting views among evaluator agents. The second error happens when some agents will be influenced by other agents and deviate from their correct initial assessments. 
To alleviate this, \method can start with $m$ separate simultaneous debate sessions ($m$ sessions similar to the session shown in Figure \ref{fig:overview}), each with the same number of agents. The sessions will continue independently to reach a final label. Once all sessions are over, the final label can be generated by aggregation over the responses from different sessions.
Having multiple independent debate sessions can help with the overall performance as any error in assessing the summary in one of the sessions will not be propagated to other sessions.
The aggregation can be done in two ways: \emph{debate vote} -- the majority vote over labels assigned in debates. Each debate session concludes with a label as described in the single debate setting.
The majority vote over these values is the final faithfulness label -- and \emph{agent vote} -- the majority vote over all participating agents in all debates. Regardless of the session to which agents belong, their individual responses are aggregated (with a majority vote) and reported as the final label.

This setup can be seen as having more evaluator agents to perform the same task, except that since sessions are independent, if there is an error propagation in one of the sessions, it will only affect the output of that session which would hopefully not affect the final aggregated response.
Also, having more agents can increase the context size (specially in the final rounds) which might not be feasible given the context size limits of some LLMs.

\section{Defining and Annotating Ambiguity}\label{sec:ambiguity}
\begin{table*}
\small
\centering
\begin{tabular}{@{}p{2.1cm}p{3cm}p{5.8cm}p{3.5cm}@{}}
\toprule
\multicolumn{1}{c}{\multirow{2}{*}{\textbf{Category}}} & \multicolumn{1}{c}{\multirow{2}{*}{\textbf{Definition}}}                                                   & \multicolumn{2}{c}{\textbf{Example}}    
\\
\cmidrule{3-4}
\multicolumn{1}{c}{}                                   & \multicolumn{1}{c}{}                                                                                       & \multicolumn{1}{c}{\textbf{Document}} & \multicolumn{1}{c}{\textbf{Summary}} \\
\midrule
Implicit reasoning phenomena                           & There is an implicit inference in the summary that can not be directly traced back to the source document. &          …The boy \textcolor{teal}{was rescued by his parents from the pit} before firefighters and paramedics arrived on the scene…
                             &                            his parents \textcolor{violet}{jumped in and pulled him to safety} before paramedics arrived.
          \\
\midrule
Meaning phenomena                                      & Summary can imply different meanings and parses.                                                           &             The 56ft (17.1m) \textcolor{teal}{converted trawler} was 6 miles (10 km) west of South Stack when the crew radioed coastguards at 07:00 BST…
                          &                   A lifeboat has been launched after a \textcolor{violet}{fishing boat} started taking on water off Anglesey.
                   \\
\midrule
Context phenomena                                      & Summary describes something correct but out of context or in a different context.                                            &              …After weighing the evidence, experts say there is a clear therapeutic role for medical cannabis. \textcolor{teal}{There is good evidence that it helps alleviate the symptoms of chronic pain, MS and nausea associated with chemotherapy, as well as anxiety.} \textcolor{orange}{But for treating other conditions, such as depression, headaches and epilepsy, there is limited or no convincing evidence that it works…}
                         &             A group of MPs has called on the government to legalize medical cannabis after a study found that one million people across the UK rely on the drug for medical reasons, but \textcolor{violet}{there is limited or no convincing evidence that it works.}     
\\
\bottomrule
\end{tabular}
\vspace*{-0.2cm}
\caption{High-level ambiguity taxonomy with definitions and color-coded exemplars. Example 1: ``jumped in'' can not be directly inferred from the document. Example 2: ``fishing boat'' can have a different meaning from ``converted trawler''. Example 3: the highlighted part in the summary can trace back to two options (colored) in the documents. The full taxonomy table can be found in the Appendix \ref{app:taxonomy}.}
\label{tab:coarse-ambiguity}
\vspace*{-0.3cm}
\end{table*}

\begin{figure}
    \centering
    \includegraphics[width=\linewidth]{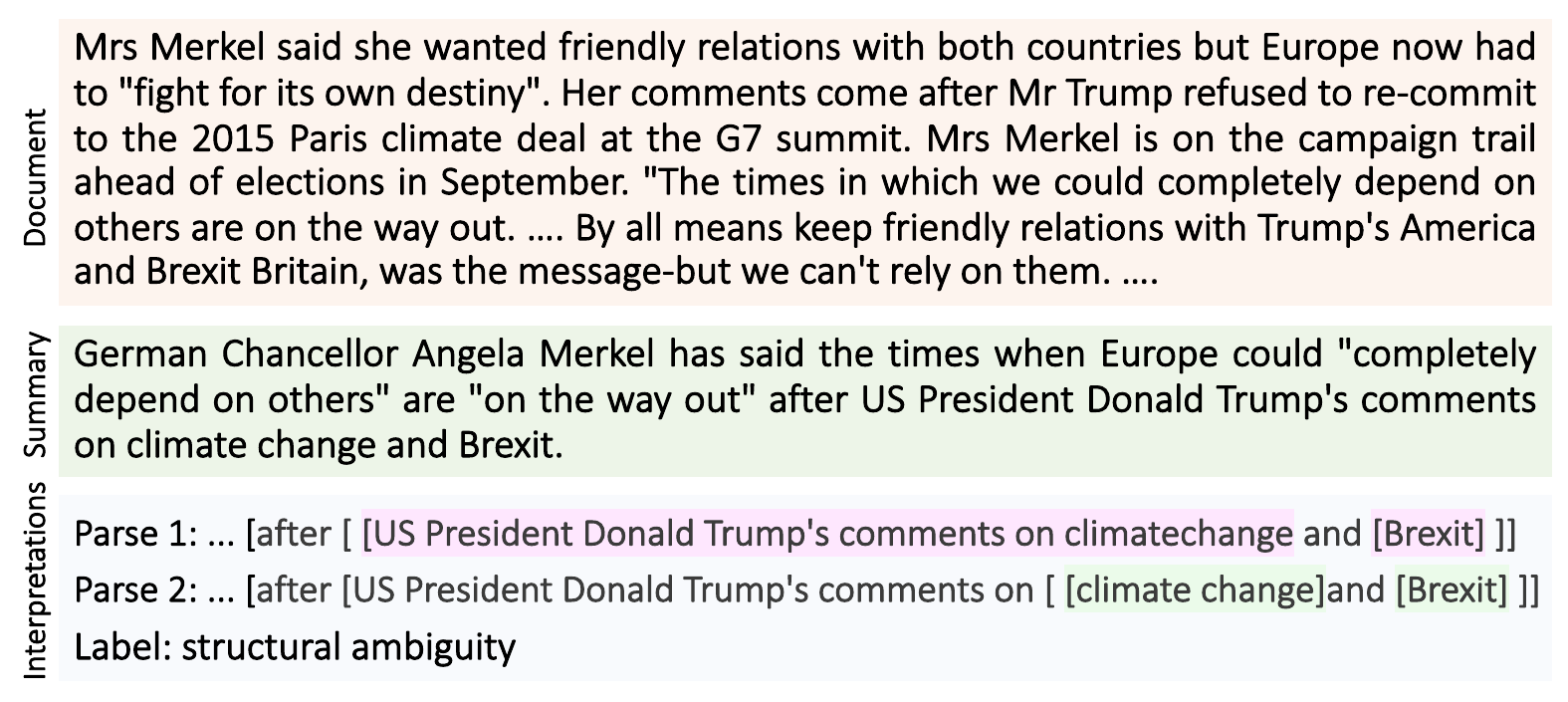}
    \vspace*{-0.7cm}
    \caption{A summary with structural ambiguity which can be interpreted using two different parses where interpretation with Parse 1 makes the summary faithful whereas Parse 2 makes the summary unfaithful.}
    \label{fig:ambiguity-motiv}
    \vspace*{-0.4cm}
\end{figure}

Faithfulness evaluation is usually done with a major underlying assumption: the summaries can \textsc{always} be definitely classified as either faithful or with some faithfulness errors. However this might not always be true. A summary can be interpreted in different ways, all plausible but where one interpretation can make the summary faithful whereas with a different interpretation, one might consider the summary as unfaithful. Given the example in Figure \ref{fig:ambiguity-motiv}, depending on how one would parse the summary, two interpretations can emerge; the first one would make the summary faithful but the second one would give the unfaithfulness perception.

We therefore define the notion of \textit{ambiguity} as follows:
an ambiguous summary can be correctly interpreted in multiple ways given the source document, leading to different faithfulness judgments depending on the underlying assumptions.
The first point about this definition is that we define ambiguity \textit{in context of the given document}. That means a summary is considered ambiguous if it can have different interpretations with respect to the source document and not on its own.
The first point is necessary but not sufficient. The sufficient condition is stated in the second point of the definition which specifies that different interpretations should lead to \textit{different faithfulness judgments} for a summary to be considered ambiguous. 
We argue that this ambiguity dimension plays a critical role in our understandings of faithfulness of the generated summaries and believe that it should be addressed before evaluating the summaries faithfulness so that evaluators would not be penalized solely based on the subjectivity of their interpretations. 
An ideal faithfulness evaluation framework should hence include an ambiguity detection module to filter out the ambiguous cases and perform faithfulness evaluation on non-ambiguous instances only (we depict an example under our multi-agent debate framework in the appendix Figure~\ref{fig:overview_w_ambiguity}).

To better help with identifying ambiguous cases as defined above, we first introduce a detailed taxonomy of such cases along with the definitions and examples of each category in Section \ref{sec:taxonomy}. Then, in a first attempt to identify ambiguous cases in summaries, we extend TofuEval MeetingBank \cite{tang2024tofueval} with ambiguity human annotations and present the details in Section \ref{sec:data_annotation}.

\subsection{Ambiguity Taxonomy}\label{sec:taxonomy}
We used our definition of ambiguity as stated above and tried to classify the ambiguities into respective categories. We have looked into possible causes of ambiguity and come up with a fine-grained taxonomy that consists of $16$ different ambiguity types. 
On a coarse level, ambiguities can be grouped into three main categories summarized as follows (see examples in Table \ref{tab:coarse-ambiguity}):

\noindent\textbf{Implicit reasoning phenomena.} This category refers to summary instances containing some type of implicit reasoning that can not be directly traced back to the document which would lead to difficulty in evaluating the summary faithfulness. The main sub-categories are deduction and inference.

\noindent\textbf{Meaning phenomena.}
This includes cases where there are multiple meanings associated with the summary which makes it ambiguous. The meaning phenomena can cover different semantic relations, linguistic ambiguity or vagueness. 

\noindent\textbf{Context phenomena.}
This category deals with summaries that are ambiguous as a result of challenges of representing the information of the source document as part of the summary. It includes decontextualization and conflation as the two main sub-categories.

The full taxonomy with fine-grained types and definitions and also a complete list of examples of each category can be found in Appendix \ref{app:taxonomy}.



\subsection{Data Annotation}\label{sec:data_annotation}
Our ambiguity benchmark is constructed on top of the TofuEval MeetingBank \cite{tang2024tofueval} faithfulness dataset. 
Professional linguists as annotators are given a detailed instructions of the task (Appendix \ref{app:ambiguity-data-stats}), its goal and the desired output. 
Next, they are provided with the document and the summary sentence and are asked to identify whether there is an ambiguity in the summary that would affect its evaluability and if so, what is the best category to describe the ambiguity using the fine-grained taxonomy in Table \ref{tab:ambiguity2}. They are also asked to write a description of the evaluability issue within the summary sentence.

Due to the inherent difficulty of the task and to ensure high inter-annotator agreement, we performed a final step to finalize the ambiguity annotations. For each instance, two experts (well-familiar with the taxonomy and the task) went over the responses by both annotators and made the final call on whether there is an ambiguity or not and if so, picked the best category from the taxonomy. 
The data statistics is shown in Table \ref{tab:ambiguity_stats}. The final dataset has an inter-annotator agreement (Cohen's Kappa) of $\approx0.73$ on binary labels.
More on IAA and the distribution of fine-grained sub-categories can be found in Appendix \ref{app:ambiguity-data-stats} and Table \ref{tab:ambiguity_stats_fine}.



\subsection{Ambiguity Detection}
Ambiguities as described earlier can lead to different assessments of faithfulness and should be addressed before evaluating the summaries so that models would not be penalized solely based on the subjectivity of the interpretations.
But how can we identify such ambiguities?
We propose an ambiguity detection approach based on \method, in which an ambiguity detector model would make a judgment call based on the arguments generated during debate. 
Formally, an ambiguity detector model predicts whether a summary sentence is ambiguous or not given the source document.
\[M_{a}(D,s,A,t) \in \text{\{ambiguous, non-ambiguous\}}\]
Where $D$ is the source document, $s$ is the summary sentence, $A$ is the arguments from agents involved in faithfulness evaluation in \method and $t$ is our proposed ambiguity taxonomy in Section \ref{sec:taxonomy}.
The overview of the full faithfulness evaluation pipeline with ambiguity detection module is shown in Figure \ref{fig:overview_w_ambiguity}. 
Evaluator agents start with opposing views on the faithfulness of the summary and try to come up with arguments to support their decisions in multiple rounds of debate. Agents with different stances can have plausible arguments for their decisions showing the possibility of an inherent ambiguity in the summary. Therefore, our proposed ambiguity detection approach makes use of the generated arguments and check their plausibility to help with understanding the ambiguities as follows: if there are sound arguments both supporting the faithfulness of the summary as well as some sound arguments arguing for the unfaithfulness of the summary sentence, the summary will be deemed ambiguous by the ambiguity detector module.
We later show how the presence of agents debate arguments can help with better identifying existing ambiguities in the summaries.

\begin{table}
\centering
\begin{tabular}{@{}l|c@{}}
\toprule
Dataset                       & MeetingBank \\ \midrule
Annotated sentences          & 770         \\
Identified ambiguous sentences & 131         \\ \midrule
Implicit reasoning ambiguities & 29\% \\
Meaning ambiguities & 29\% \\
Context ambiguities & 34\% \\
Other ambiguities & 7\% \\
\bottomrule
\end{tabular}
\vspace*{-0.2cm}
\caption{Statistics of MeetingBank dataset annotated for ambiguity along with the distribution of high-level categories. Fine-grained distribution in Table \ref{tab:ambiguity_stats_fine}.}
\label{tab:ambiguity_stats}
\vspace*{-0.0cm}
\end{table}
\section{Experimental Setting}
\begin{table*}
\centering
\small
\begin{tabular}{
    l
    S[table-format = 2.1] c
    S[table-format = 2.1] c
    S[table-format = 2.1] c
    S[table-format = 2.1] c
}
\toprule
\multirow{3}{*}{\textbf{Model}} & \multicolumn{4}{c}{\textbf{TofuEval}}                                            & \multicolumn{4}{c}{\textbf{AggreFact}}                                 \\ \cmidrule(l){2-5} \cmidrule(l){6-9} 
                                & \multicolumn{2}{c}{\textbf{MeetingBank}} & \multicolumn{2}{c}{\textbf{MediaSum}} & \multicolumn{2}{c}{\textbf{CNN}}   & \multicolumn{2}{c}{\textbf{XSum}} \\
                                \cmidrule(r){2-3}\cmidrule(r){4-5} \cmidrule(r){6-7} \cmidrule(r){8-9}
                                & \multicolumn{1}{l}{BAcc}    & K-alpha    & BAcc             & K-alpha            & \multicolumn{1}{l}{BAcc} & K-alpha & BAcc           & K-alpha        

\\ \cmidrule(r){1-9}
Zero-shot single LLM& 68.15	& 0.38 & 56.23	& 0.00 & 60.18 &	0.28	&68.13&	0.35
 \\
Zero-shot Chain of Thought&68.45 &	0.39	& 58.77	& 0.09 &	63.34	& 0.35 &	68.17 &	0.35
\\
Self-consistency &69.05	& 0.40 &	61.07	& 0.15	& 62.56 &	0.34&	68.87&	0.37
\\
\cmidrule(r){1-9}
\method wo initialization & 69.13	& 0.40 &	63.13&	0.20&	60.30&	0.28	& 70.18	& 0.38
\\
\method & 75.08	& 0.50	& 66.57 &	0.33 &	66.88	& 0.34 &	\bfseries 75.10 &	\bfseries 0.50
\\
\cmidrule(r){1-9}
\method w. simul. debates (agents vote) & \bfseries 78.06	& \bfseries 0.57 	& 69.25 &	0.39	& \bfseries 69.13	& \bfseries 0.39 &	73.62	& 0.47
 \\
\method w. simul. debates (debates vote) & 	77.42	& 0.56 &	\bfseries 70.59 &	\bfseries 0.42 & 69.03	& 0.39	& 74.71 & 	0.49
 \\
\bottomrule
\end{tabular}
\vspace*{-0.2cm}
\caption{Results of different faithfulness evaluators. The first three are our baselines, while the last four are the variants of \method. The best results for each dataset are highlighted. A more detailed comparison with other evaluators are presented in Table \ref{tab:full}.}
\label{tab:llama-main}
\vspace*{-0.2cm}
\end{table*}

\subsection{Datasets}
To evaluate our multi agent debate framework \method, we use a mix of summarization datasets, namely AggreFact \cite{tang-etal-2023-understanding} benchmark consisting of CNN and XSum datasets 
as well as TofuEval benchmark \cite{tang2024tofueval} consisting of an annotated subset of MediaSum \cite{zhu2021mediasum} and MeetingBank \cite{hu2023meetingbank}, for a mix of news and dialogue domains. The ambiguity annotation of MeetingBank (Section~\ref{sec:ambiguity} is additionaly used for ambiguity related experiments. 
The statistics of the datasets are presented in Table \ref{tab:dataset}.
We have used full summaries (instead of sentence-level) to measure faithfulness on TofuEval, as it was previously shown that asking the model to evaluate sentences at once or individually would not lead to any significant performance change \cite{tang2024minicheck}. However, we also report the sentence-level results in Appendix \ref{app:result}.

\subsection{Evaluators}
We use Meta Llama3 \cite{llama3modelcard} as our underlying LLM for our experiments and results reported in the main script. We also used other LLMs and reported the results in Appendix \ref{app:result}.
We have used different setups, including single and multi-LLM evaluators and compared their performance with variations of \method:
%
\textbf{(1) Zero-shot Single LLM:} a single LLM agent which is directly asked to predict whether the given summary is faithful or not given the document. 
%
\textbf{(2) Chain of thought:} an LLM is asked to first think step by step before providing its judgment on the summary \cite{wei2022chain}.
%
\textbf{(3) Self-consistency:} the system is queried $n$ times \cite{wang2022self} to sample different paths, with the final judgment determined by the majority vote.
%
\textbf{(4) \method wo. initialization:} 
\method with $4$ evaluator agents participating in at most $3$ discussion rounds to evaluate the faithfulness of the summary as shown in Figure \ref{fig:overview} but without the stance initialization stage. 
%
\textbf{(5) \method:} our proposed approach and evaluation framework as shown in Figure \ref{fig:overview} with $4$ evaluator agents and at most $3$ discussion rounds.
\textbf{(6) \method w. simultaneous debates: }
instead of having a single debate session, we initialize $3$ simultaneous debate sessions, each with $4$ evaluator agents, and the final label would be aggregated over the responses from different sessions as described in \ref{app:method}.
%
All setups perform the evaluation in a zero-shot manner.
The prompts used for all these settings are presented in Appendix \ref{app:prompts}.

\subsection{Evaluation Criteria}
We have used two main metrics for our evaluation purposes, balanced accuracy (BAcc) which is used to measure the overall performance of evaluators in detecting the correct labels for summaries, and Krippendorff alpha (K-alpha) \cite{krippendorff2011computing} to measure how well system-generated labels align with the human annotations. More details on these metrics can be found in Appendix \ref{app:metrics}.

\section{Evaluation}

Our evaluation setup is focused on three main directions; First, 
showing the improvement of \method in terms of accuracy for faithfulness evaluation plus the added interpretability with generated explanations for faithfulness label.
Second, justifying our arguments on how ambiguity can affect the performance of faithfulness evaluators and how addressing them can help with better assessment of performance.
Finally, showing that \method does not only help with better faithfulness evaluation but it also helps with identifying ambiguity.

\paragraph{How does \method compare with other single and multi LLM-based baselines?}
We report BAcc and K-alpha of different models using Llama3-70B-instruct in Table \ref{tab:llama-main}.
Overall, \method improves performance on faithfulness evaluation task compared to all other baselines, and the predictions are better aligned with human annotations.
Moreover, our approach is orthogonal to the underlying LLM and we also observe similar trends for other LLMs as well (Appendix \ref{app:results:gpt4}, \ref{app:results:llama-small}). 
For a more complete set of results, both sentence-level and summary-level using different automatic evaluators, refer to Table \ref{tab:full} in Appendix \ref{app:full_results}.

\paragraph{How effective is the initial stance assignment?}
One of the key components of \method is the stance initialization stage where the evaluator agents are assigned opposing beliefs about the faithfulness of the summary before entering the debate stage as shown in Figure \ref{fig:overview}. 
Assigning initial stances to evaluator agents can significantly improve the performance of \method as this initialization encourages LLMs to think more thoroughly as to whether there exists a faithfulness error in the summary or not. As shown in Table \ref{tab:llama-main}, \method without initialization performs almost similarly to other baselines. But after assigning the random stances, a larger performance gap is observed as shown in the second chunk of Table \ref{tab:llama-main}, highlighting the importance of initialization to diversify the debate towards identifying more errors (for analysis on the effect of stance initialization distribution, please refer to Appendix \ref{app:init}).

\paragraph{Can \method identify more errors?}
Missing on a large portion of the errors in the summaries is a major issue with the existing evaluation approaches. This mainly happens due to the fact that evaluators are usually fooled by the fluency of the generated text and would fail to distinguish fluency from faithfulness. This might be even more problematic in domains where failure to identify an error in the text can be a critical issue (for instance medical domain).
We report the false negative rate (FNR) and false positive rate (FPR) as described in Appendix \ref{app:metrics} in Table \ref{tab:llama-main-fpr-fnr}.
It is shown that \method is capable of achieving lower FNR by identifying more errors with the help of random stance initialization and debate.
However, since \method is more sensitive to the errors, the FPR is also increased. 
More on  why this might be the case and how it can be alleviated is described in Appendix \ref{app:results:fpr-fnr}.

\paragraph{Can \method help identify ambiguities?}
Ambiguities as described earlier can lead to different assessments of faithfulness and should be addressed before evaluating the summaries so that models would not be penalized solely based on the subjectivity of the interpretations. 
\begin{table}
\centering
\begin{tabular}{l|S[table-format = 2.1]}
\toprule
\textbf{Model}  & \textbf{BAcc}
\\ 
\midrule
Random baseline & 50.00
\\
self-consistency variation & 52.00
\\
Baseline w ambiguity taxonomy & 59.26
\\
Debate disagreement& 66.09
\\
Debate arguments & \bfseries 71.37 \\
\bottomrule
\end{tabular}
\caption{Ambiguity detection balanced accuracy. The arguments generated using \method can help with identifying ambiguous cases.}
\label{tab:ambiguity_results}
\end{table}


But how can we identify such ambiguities? Using our proposed taxonomy and the MeetingBank annotated data on this dimension (as described in Section \ref{sec:data_annotation}), we tried different ways to automatically identify such cases given the document and the summary using Llama3-70B-instruct:
\textbf{1. Self-consistency variation:} 
In this baseline, LLM is asked multiple times ($41$ in our case) to identify whether the summary is faithful or not. Then, the ratio of the times the answer is faithful and the ratio of the times the summary is labeled as unfaithful will be measured. If the difference lies between some pre-define threshold ($<20$), the summary will be considered as ambiguous. The motivation using this approach is that if the evaluator is not sure of its decision, that can mean the summary can be interpreted in different ways, hence ambiguous.
\textbf{2. Zero-shot with ambiguity taxonomy:}
We provide our ambiguity taxonomy to LLM to identify whether the summary is ambiguous or not.
\textbf{3. Debate disagreement:}
Using \method, we consider cases for which even after $3$ rounds of debate, none of the agents changed their initial stances as ambiguous.
\textbf{4. Ambiguity detection with debate arguments:}
Using the arguments of the debates and ambiguity taxonomy, we ask the LLM to identify whether there exists an ambiguity or not. You can refer to prompts in Table \ref{tab:prompt_lists} in Appendix \ref{app:prompts}.

\noindent 
The accuracy numbers are reported in Table \ref{tab:ambiguity_results}. 
The ambiguity taxonomy can help baselines with identifying the ambiguous cases. Our best performing ambiguity detection model is the one which uses the arguments from the debates on summary faithfulness. 
Our results suggest that not only does \method help with faithfulness evaluation but it can also serve as a means to identifying ambiguous cases and filtering them. 
These are the initial results on ambiguity detection however there is still a large room for improvement on the task which is left for future work.

\begin{figure}
\large
\begin{minipage}{0.23\textwidth} \centering
\begin{tikzpicture}[scale=0.45]
    \begin{axis}[
       ybar=2*\pgflinewidth,
        bar width=0.8cm,
        ymajorgrids = true,
        grid style=dashed,
        ylabel={BAcc},
        nodes near coords,
        symbolic x coords={Zero-shot,Consistency,Debate},
        xtick = data,
        scaled y ticks = false,
        enlarge x limits=0.25,
        ymin=50,
        ymax=95,
        legend columns=2,
        legend style={at={(0.05,-0.15)},anchor=north west},
    ]

    \addplot[style={black,pattern color=black,pattern = north west lines}]
    coordinates {(Zero-shot,75.57)(Consistency,74.71)(Debate,79.67)};
    
    \addplot[style={black,pattern color=teal,pattern = crosshatch}]
    coordinates {(Zero-shot,82.59)(Consistency,80.99)(Debate,87.53)};

\addlegendentry{Unfiltered}
\addlegendentry{Filtered}

    \end{axis}
\end{tikzpicture}
\end{minipage}\hspace{0.1em}
\begin{minipage}{0.23\textwidth} \centering
\begin{tikzpicture}[scale=0.45]
    \begin{axis}[
       ybar=2*\pgflinewidth,
        bar width=0.8cm,
        ymajorgrids = true,
        grid style=dashed,
        ylabel={K-alpha},
        nodes near coords,
        symbolic x coords={Zero-shot,Consistency,Debate},
        xtick = data,
        scaled y ticks = false,
        enlarge x limits=0.25,
        ymin=0.2,
        ymax=0.8,
        legend columns=2,
        legend style={at={(0.05,-0.15)},anchor=north west},
    ]

    \addplot[style={black,pattern color=black,pattern = north west lines}]
    coordinates {(Zero-shot,0.52)(Consistency,0.52)(Debate,0.53)};
    
    \addplot[style={black,pattern color=teal,pattern = crosshatch}]
    coordinates {(Zero-shot,0.67)(Consistency,0.65)(Debate,0.71)};

\addlegendentry{Unfiltered}
\addlegendentry{Filtered}
\end{axis}
\end{tikzpicture}
\end{minipage}\hfill

\caption{BAcc and correlation to human judgements results pre (black) and post (teal) filtering the ambiguous cases on annotated MeetingBank dataset.}
  \label{plot:ambiguity-filtering}
\end{figure}
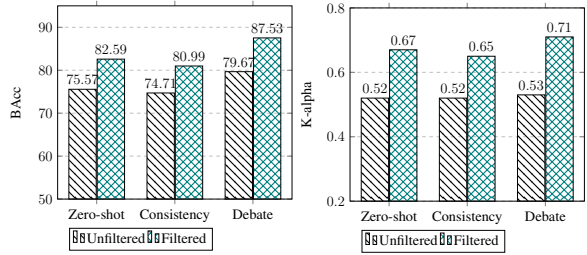
\paragraph{How ambiguous cases can affect the evaluators performance?}
As can be seen from Table \ref{tab:llama-main}, even the best performing evaluators still fall very short in terms of k-alpha showing low agreement between models predictions and human annotations. Aside from the evaluators individual errors, the existence of ambiguities is a major contributing factor to low agreement and would lead to incorrect conclusions on models performance.

To remove the effect of ambiguous cases on model performance and have a more accurate estimate of evaluators performance, we filtered them out (the ones annotated as ambiguous by human annotators) from the evaluation subset (MeetingBank dataset with ambiguity annotation) and measured the performance of different models on both unfiltered/filtered data. 
As can be seen in Figure \ref{plot:ambiguity-filtering}, regardless of the setting, removing such ambiguous cases would lead to higher agreement between gold labels and the model-generated labels (with slightly larger gap for \method). Removing ambiguities can also improve FNR and FPR trends (\ref{app:results:fpr-fnr-filtered}).



\section{Conclusion}
We have proposed \method, a new automatic LLM-based multi-agent summary faithfulness evaluation with stance initialization and multi-round debate shown to be capable of identifying more errors compared to other LLM-based baselines. We have also identified a new evaluation dimension called \textit{ambiguity} and a detailed taxonomy to identify ambiguous summaries that can be evaluated as both faithful and unfaithful depending on the how one would interpret them. We extend the MeetingBank dataset by providing annotations for ambiguity dimension and show how filtering the ambiguous cases can help further improve the results and lead to higher IAA.
\section*{Limitations}
Our work has some limitations.
First, we have not used a large set of LLMs for our experiments as the primary goal of our work was to show the relative improvement of \method compared to other baseline settings with a specific LLM and how this approach can help with faithfulness evaluation regardless of the underlying LLM.
Second, our faithfulness evaluation is aimed at generating a final binary label for the non-ambiguous summaries for our choice of datasets. However, \method can be modified to ask for a faithfulness rating rather than a binary label. This can further improve the evaluation of summarizers on a finer-grained level. This can be a direction for future work.
Finally, ambiguity annotation is only done on sentence-level. More analysis is required to see whether ambiguities can span over a sentence.

\bibliography{custom,anthology}

\begin{thebibliography}{36}
\providecommand{\natexlab}[1]{#1}

\bibitem[{AI@Meta(2024)}]{llama3modelcard}
AI@Meta. 2024.
\newblock \href {https://github.com/meta-llama/llama3/blob/main/MODEL_CARD.md} {Llama 3 model card}.

\bibitem[{Chan et~al.(2023)Chan, Chen, Su, Yu, Xue, Zhang, Fu, and Liu}]{chan2023chateval}
Chi-Min Chan, Weize Chen, Yusheng Su, Jianxuan Yu, Wei Xue, Shanghang Zhang, Jie Fu, and Zhiyuan Liu. 2023.
\newblock Chateval: Towards better llm-based evaluators through multi-agent debate.
\newblock \emph{arXiv preprint arXiv:2308.07201}.

\bibitem[{Chen et~al.(2019)Chen, Bragg, Chilton, and Weld}]{chen2019cicero}
Quanze Chen, Jonathan Bragg, Lydia~B Chilton, and Dan~S Weld. 2019.
\newblock Cicero: Multi-turn, contextual argumentation for accurate crowdsourcing.
\newblock In \emph{Proceedings of the 2019 chi conference on human factors in computing systems}, pages 1--14.

\bibitem[{Clark et~al.(2023)Clark, Rijhwani, Gehrmann, Maynez, Aharoni, Nikolaev, Sellam, Siddhant, Das, and Parikh}]{clark2023seahorse}
Elizabeth Clark, Shruti Rijhwani, Sebastian Gehrmann, Joshua Maynez, Roee Aharoni, Vitaly Nikolaev, Thibault Sellam, Aditya Siddhant, Dipanjan Das, and Ankur Parikh. 2023.
\newblock Seahorse: A multilingual, multifaceted dataset for summarization evaluation.
\newblock In \emph{Proceedings of the 2023 Conference on Empirical Methods in Natural Language Processing}, pages 9397--9413.

\bibitem[{Du et~al.(2023)Du, Li, Torralba, Tenenbaum, and Mordatch}]{du2023improving}
Yilun Du, Shuang Li, Antonio Torralba, Joshua~B Tenenbaum, and Igor Mordatch. 2023.
\newblock Improving factuality and reasoning in language models through multiagent debate.
\newblock \emph{arXiv preprint arXiv:2305.14325}.

\bibitem[{Elangovan et~al.(2024)Elangovan, Liu, Xu, Bodapati, and Roth}]{elangovan2024considersthehumanevaluationframeworkrethinking}
Aparna Elangovan, Ling Liu, Lei Xu, Sravan Bodapati, and Dan Roth. 2024.
\newblock \href {https://arxiv.org/abs/2405.18638} {Considers-the-human evaluation framework: Rethinking human evaluation for generative large language models}.
\newblock \emph{Preprint}, arXiv:2405.18638.

\bibitem[{Fabbri et~al.(2022)Fabbri, Wu, Liu, and Xiong}]{fabbri2022qafacteval}
Alexander~Richard Fabbri, Chien-Sheng Wu, Wenhao Liu, and Caiming Xiong. 2022.
\newblock Qafacteval: Improved qa-based factual consistency evaluation for summarization.
\newblock In \emph{Proceedings of the 2022 Conference of the North American Chapter of the Association for Computational Linguistics: Human Language Technologies}, pages 2587--2601.

\bibitem[{Gao and Wan(2022)}]{gao-wan-2022-dialsummeval}
Mingqi Gao and Xiaojun Wan. 2022.
\newblock \href {https://doi.org/10.18653/v1/2022.naacl-main.418} {{D}ial{S}umm{E}val: Revisiting summarization evaluation for dialogues}.
\newblock In \emph{Proceedings of the 2022 Conference of the North American Chapter of the Association for Computational Linguistics: Human Language Technologies}, pages 5693--5709, Seattle, United States. Association for Computational Linguistics.

\bibitem[{Goyal and Durrett(2020)}]{goyal2020evaluating}
Tanya Goyal and Greg Durrett. 2020.
\newblock Evaluating factuality in generation with dependency-level entailment.
\newblock In \emph{Findings of the Association for Computational Linguistics: EMNLP 2020}, pages 3592--3603.

\bibitem[{Hong et~al.(2023)Hong, Zheng, Chen, Cheng, Wang, Zhang, Wang, Yau, Lin, Zhou et~al.}]{hong2023metagpt}
Sirui Hong, Xiawu Zheng, Jonathan Chen, Yuheng Cheng, Jinlin Wang, Ceyao Zhang, Zili Wang, Steven Ka~Shing Yau, Zijuan Lin, Liyang Zhou, et~al. 2023.
\newblock Metagpt: Meta programming for multi-agent collaborative framework.
\newblock \emph{arXiv preprint arXiv:2308.00352}.

\bibitem[{Hu et~al.(2023)Hu, Ganter, Deilamsalehy, Dernoncourt, Foroosh, and Liu}]{hu2023meetingbank}
Yebowen Hu, Tim Ganter, Hanieh Deilamsalehy, Franck Dernoncourt, Hassan Foroosh, and Fei Liu. 2023.
\newblock Meetingbank: A benchmark dataset for meeting summarization.
\newblock \emph{arXiv preprint arXiv:2305.17529}.

\bibitem[{Krippendorff(2011)}]{krippendorff2011computing}
Klaus Krippendorff. 2011.
\newblock Computing krippendorff’s alpha-reliability.

\bibitem[{Kry{\'s}ci{\'n}ski et~al.(2020)Kry{\'s}ci{\'n}ski, McCann, Xiong, and Socher}]{kryscinski2020evaluating}
Wojciech Kry{\'s}ci{\'n}ski, Bryan McCann, Caiming Xiong, and Richard Socher. 2020.
\newblock Evaluating the factual consistency of abstractive text summarization.
\newblock In \emph{Proceedings of the 2020 Conference on Empirical Methods in Natural Language Processing (EMNLP)}, pages 9332--9346.

\bibitem[{Lan et~al.(2024)Lan, Gao, Jin, and Li}]{lan2024stance}
Xiaochong Lan, Chen Gao, Depeng Jin, and Yong Li. 2024.
\newblock Stance detection with collaborative role-infused llm-based agents.
\newblock In \emph{Proceedings of the International AAAI Conference on Web and Social Media}, volume~18, pages 891--903.

\bibitem[{Liang et~al.(2023)Liang, He, Jiao, Wang, Wang, Wang, Yang, Tu, and Shi}]{DBLP:journals/corr/abs-2305-19118}
Tian Liang, Zhiwei He, Wenxiang Jiao, Xing Wang, Yan Wang, Rui Wang, Yujiu Yang, Zhaopeng Tu, and Shuming Shi. 2023.
\newblock \href {https://doi.org/10.48550/ARXIV.2305.19118} {Encouraging divergent thinking in large language models through multi-agent debate}.
\newblock \emph{CoRR}, abs/2305.19118.

\bibitem[{Lin(2004)}]{lin2004rouge}
Chin-Yew Lin. 2004.
\newblock Rouge: A package for automatic evaluation of summaries.
\newblock In \emph{Text summarization branches out}, pages 74--81.

\bibitem[{Liu et~al.(2023)Liu, Iter, Xu, Wang, Xu, and Zhu}]{liu2023g}
Yang Liu, Dan Iter, Yichong Xu, Shuohang Wang, Ruochen Xu, and Chenguang Zhu. 2023.
\newblock G-eval: Nlg evaluation using gpt-4 with better human alignment.
\newblock \emph{arXiv preprint arXiv:2303.16634}.

\bibitem[{Luo et~al.(2023)Luo, Xie, and Ananiadou}]{luo2023chatgptfactualinconsistencyevaluator}
Zheheng Luo, Qianqian Xie, and Sophia Ananiadou. 2023.
\newblock \href {https://arxiv.org/abs/2303.15621} {Chatgpt as a factual inconsistency evaluator for text summarization}.
\newblock \emph{Preprint}, arXiv:2303.15621.

\bibitem[{Mandi et~al.(2024)Mandi, Jain, and Song}]{mandi2024roco}
Zhao Mandi, Shreeya Jain, and Shuran Song. 2024.
\newblock Roco: Dialectic multi-robot collaboration with large language models.
\newblock In \emph{2024 IEEE International Conference on Robotics and Automation (ICRA)}, pages 286--299. IEEE.

\bibitem[{Maynez et~al.(2020)Maynez, Narayan, Bohnet, and McDonald}]{maynez-etal-2020-faithfulness}
Joshua Maynez, Shashi Narayan, Bernd Bohnet, and Ryan McDonald. 2020.
\newblock \href {https://doi.org/10.18653/v1/2020.acl-main.173} {On faithfulness and factuality in abstractive summarization}.
\newblock In \emph{Proceedings of the 58th Annual Meeting of the Association for Computational Linguistics}, pages 1906--1919, Online. Association for Computational Linguistics.

\bibitem[{Papineni et~al.(2002)Papineni, Roukos, Ward, and Zhu}]{papineni2002bleu}
Kishore Papineni, Salim Roukos, Todd Ward, and Wei-Jing Zhu. 2002.
\newblock Bleu: a method for automatic evaluation of machine translation.
\newblock In \emph{Proceedings of the 40th annual meeting of the Association for Computational Linguistics}, pages 311--318.

\bibitem[{Qian et~al.(2024)Qian, Liu, Liu, Chen, Dang, Li, Yang, Chen, Su, Cong et~al.}]{qian2024chatdev}
Chen Qian, Wei Liu, Hongzhang Liu, Nuo Chen, Yufan Dang, Jiahao Li, Cheng Yang, Weize Chen, Yusheng Su, Xin Cong, et~al. 2024.
\newblock Chatdev: Communicative agents for software development.
\newblock In \emph{Proceedings of the 62nd Annual Meeting of the Association for Computational Linguistics (Volume 1: Long Papers)}, pages 15174--15186.

\bibitem[{Smit et~al.(2024)Smit, Grinsztajn, Duckworth, Barrett, and Pretorius}]{DBLP:conf/icml/SmitGDBP24}
Andries~P. Smit, Nathan Grinsztajn, Paul Duckworth, Thomas~D. Barrett, and Arnu Pretorius. 2024.
\newblock \href {https://openreview.net/forum?id=CrUmgUaAQp} {Should we be going mad? {A} look at multi-agent debate strategies for llms}.
\newblock In \emph{Forty-first International Conference on Machine Learning, {ICML} 2024, Vienna, Austria, July 21-27, 2024}. OpenReview.net.

\bibitem[{Song et~al.(2024)Song, Su, Shalyminov, Cai, and Mansour}]{song2024finesure}
Hwanjun Song, Hang Su, Igor Shalyminov, Jason Cai, and Saab Mansour. 2024.
\newblock Finesure: Fine-grained summarization evaluation using llms.
\newblock \emph{arXiv preprint arXiv:2407.00908}.

\bibitem[{Tang et~al.(2023{\natexlab{a}})Tang, Goyal, Fabbri, Laban, Xu, Yavuz, Kryscinski, Rousseau, and Durrett}]{tang-etal-2023-understanding}
Liyan Tang, Tanya Goyal, Alex Fabbri, Philippe Laban, Jiacheng Xu, Semih Yavuz, Wojciech Kryscinski, Justin Rousseau, and Greg Durrett. 2023{\natexlab{a}}.
\newblock \href {https://doi.org/10.18653/v1/2023.acl-long.650} {Understanding factual errors in summarization: Errors, summarizers, datasets, error detectors}.
\newblock In \emph{Proceedings of the 61st Annual Meeting of the Association for Computational Linguistics (Volume 1: Long Papers)}, pages 11626--11644, Toronto, Canada. Association for Computational Linguistics.

\bibitem[{Tang et~al.(2024{\natexlab{a}})Tang, Laban, and Durrett}]{tang2024minicheck}
Liyan Tang, Philippe Laban, and Greg Durrett. 2024{\natexlab{a}}.
\newblock Minicheck: Efficient fact-checking of llms on grounding documents.
\newblock \emph{arXiv preprint arXiv:2404.10774}.

\bibitem[{Tang et~al.(2024{\natexlab{b}})Tang, Shalyminov, Wong, Burnsky, Vincent, Singh, Feng, Song, Su, Sun et~al.}]{tang2024tofueval}
Liyan Tang, Igor Shalyminov, Amy Wong, Jon Burnsky, Jake Vincent, Siffi Singh, Song Feng, Hwanjun Song, Hang Su, Lijia Sun, et~al. 2024{\natexlab{b}}.
\newblock Tofueval: Evaluating hallucinations of llms on topic-focused dialogue summarization.
\newblock In \emph{Proceedings of the 2024 Conference of the North American Chapter of the Association for Computational Linguistics: Human Language Technologies (Volume 1: Long Papers)}, pages 4455--4480.

\bibitem[{Tang et~al.(2023{\natexlab{b}})Tang, Sun, Idnay, Nestor, Soroush, Elias, Xu, Ding, Durrett, Rousseau, Weng, and Peng}]{Tang2023.04.22.23288967}
Liyan Tang, Zhaoyi Sun, Betina Idnay, Jordan~G Nestor, Ali Soroush, Pierre~A. Elias, Ziyang Xu, Ying Ding, Greg Durrett, Justin Rousseau, Chunhua Weng, and Yifan Peng. 2023{\natexlab{b}}.
\newblock \href {https://doi.org/10.1038/s41746-023-00896-7} {Evaluating large language models on medical evidence summarization}.
\newblock \emph{npj Digit. Med. 6}.

\bibitem[{Verga et~al.(2024)Verga, Hofstatter, Althammer, Su, Piktus, Arkhangorodsky, Xu, White, and Lewis}]{verga2024replacing}
Pat Verga, Sebastian Hofstatter, Sophia Althammer, Yixuan Su, Aleksandra Piktus, Arkady Arkhangorodsky, Minjie Xu, Naomi White, and Patrick Lewis. 2024.
\newblock Replacing judges with juries: Evaluating llm generations with a panel of diverse models.
\newblock \emph{arXiv preprint arXiv:2404.18796}.

\bibitem[{Wang et~al.(2024)Wang, Wang, Su, Tong, and Song}]{wang2024rethinking}
Qineng Wang, Zihao Wang, Ying Su, Hanghang Tong, and Yangqiu Song. 2024.
\newblock Rethinking the bounds of llm reasoning: Are multi-agent discussions the key?
\newblock \emph{arXiv preprint arXiv:2402.18272}.

\bibitem[{Wang et~al.(2022)Wang, Wei, Schuurmans, Le, Chi, Narang, Chowdhery, and Zhou}]{wang2022self}
Xuezhi Wang, Jason Wei, Dale Schuurmans, Quoc Le, Ed~Chi, Sharan Narang, Aakanksha Chowdhery, and Denny Zhou. 2022.
\newblock Self-consistency improves chain of thought reasoning in language models.
\newblock \emph{arXiv preprint arXiv:2203.11171}.

\bibitem[{Wang et~al.(2023)Wang, Reddy, Mujahid, Arora, Rubashevskii, Geng, Afzal, Pan, Borenstein, Pillai et~al.}]{wang2023factcheck}
Yuxia Wang, Revanth~Gangi Reddy, Zain~Muhammad Mujahid, Arnav Arora, Aleksandr Rubashevskii, Jiahui Geng, Osama~Mohammed Afzal, Liangming Pan, Nadav Borenstein, Aditya Pillai, et~al. 2023.
\newblock Factcheck-gpt: End-to-end fine-grained document-level fact-checking and correction of llm output.
\newblock \emph{arXiv preprint arXiv:2311.09000}.

\bibitem[{Wei et~al.(2022)Wei, Wang, Schuurmans, Bosma, Xia, Chi, Le, Zhou et~al.}]{wei2022chain}
Jason Wei, Xuezhi Wang, Dale Schuurmans, Maarten Bosma, Fei Xia, Ed~Chi, Quoc~V Le, Denny Zhou, et~al. 2022.
\newblock Chain-of-thought prompting elicits reasoning in large language models.
\newblock \emph{Advances in neural information processing systems}, 35:24824--24837.

\bibitem[{Zhang et~al.(2024)Zhang, Xu, Zhang, Liu, Hooi, and Deng}]{DBLP:conf/acl/ZhangX0LHD24}
Jintian Zhang, Xin Xu, Ningyu Zhang, Ruibo Liu, Bryan Hooi, and Shumin Deng. 2024.
\newblock \href {https://aclanthology.org/2024.acl-long.782} {Exploring collaboration mechanisms for {LLM} agents: {A} social psychology view}.
\newblock In \emph{Proceedings of the 62nd Annual Meeting of the Association for Computational Linguistics (Volume 1: Long Papers), {ACL} 2024, Bangkok, Thailand, August 11-16, 2024}, pages 14544--14607. Association for Computational Linguistics.

\bibitem[{Zhang et~al.(2020)Zhang, Kishore, Wu, Weinberger, and Artzi}]{zhang2020bertscoreevaluatingtextgeneration}
Tianyi Zhang, Varsha Kishore, Felix Wu, Kilian~Q. Weinberger, and Yoav Artzi. 2020.
\newblock \href {https://arxiv.org/abs/1904.09675} {Bertscore: Evaluating text generation with bert}.
\newblock \emph{Preprint}, arXiv:1904.09675.

\bibitem[{Zhu et~al.(2021)Zhu, Liu, Mei, and Zeng}]{zhu2021mediasum}
Chenguang Zhu, Yang Liu, Jie Mei, and Michael Zeng. 2021.
\newblock Mediasum: A large-scale media interview dataset for dialogue summarization.
\newblock \emph{arXiv preprint arXiv:2103.06410}.

\end{thebibliography}
\clearpage
\newpage
\appendix
\section{Multi-agent Debate Approach Details}
The following sections describe more details of our proposed approach.

\subsection{Multi-round Debate}
The multi-round debate stage of \method is \textit{guideline-based} and will be stopped it meets the \textit{stopping criterion}. We describe guidelines that are used during the debate and the stopping conditions below.

\subsubsection{Guidelines}\label{app:guidelines}
LLMs have their own interpretations of concepts and similar to human evaluators might mix their perception with what is actually considered correct or plausible \cite{elangovan2024considersthehumanevaluationframeworkrethinking} which might be different from specific needs and requirements of certain tasks. Guidelines or rules can be established to clearly specify the dos and don’ts of the evaluation process such that evaluators can base their judgment on them and can easily refer to them during discussion leading to discussion efficiency \cite{chen2019cicero}. Guidelines can encourage a more structured debate and arguments referring to the guidelines can be verified based on whether the guidelines are used correctly or not. 

Guidelines can be generated manually and provided as part of the prompt. However, it might be difficult to come up with a comprehensive set of desirable guidelines at once and prior to the evaluation. 
Instead we can apply an alternative semi-automatic (possibly automatic) approach to generate guidelines in a learning phase using the following procedure. We start with a small subset of the annotated data (both positive and negative from dev sets) and use our debate approach for the evaluation with a minor tweak. We explicitly ask agents to provide the guidelines they have used to make their judgments and collect them. Agents might be either correct or wrong in their final judgements on whether the summary is faithful or not. If an agent is correct, the guidelines provided by it will be placed in the list of potential guidelines and if it is incorrect, the negated guidelines will be added to the pool. This process is done incrementally, meaning that after each evaluation the guidelines are updated and provided to the agents. Once, no more new guidelines are added to the pool (after a certain number of repetitions), the learning phase is stopped and the full set of guidelines will be curated for future evaluations.

\begin{figure*}
    \centering
    \includegraphics[width=1\linewidth]{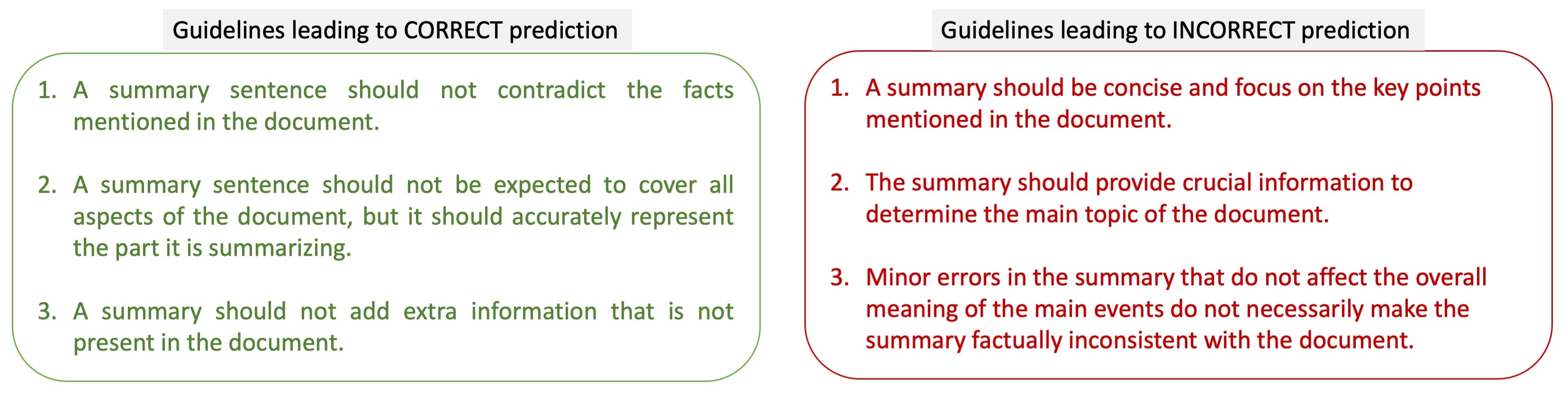}
    \caption{Guidelines generated during learning phase.}
    \label{fig:guidelines}
\end{figure*}

Figure \ref{fig:guidelines} shows some of the generated guidelines during the learning phase. Some of these guidelines lead to correct label prediction whereas the other ones can result in an incorrect prediction. The later group should be negated and provided to the agents for future predictions.

\subsubsection{Stopping Criterion}\label{app:stop}
At any debate round $r_j$ if agents reach consensus and \textbf{\textit{all}} agree on the faithfulness label, the debate would be stopped and label $l$ would be assigned to the summary. 
However, it might be the case that even after rounds of debate, there would still be disagreement among agents. In such cases, once the debate reaches its predefined maximum number of rounds $n$, the debate will be stopped and the final decision would be made in the adjudication step.

If after multiple rounds of discussions, the agents still disagree, an intervention happens and agents are encouraged to be more open to accept each other’s opinion. This can be done by either updating the description of the task that has been assigned to them or through specifying a new goal.

Finally, after a fixed number of rounds, the debate will be stopped and the final decision would be made in the adjudication step.

\subsection{Evaluation Metrics}\label{app:metrics}
\subsubsection{Balanced Accuracy}
Following previous works, we evaluate the performance of our evaluation approach using Balanced Accuracy (BAcc). This metric takes into account the imbalance of consistent and inconsistent summaries with respect to the evaluation dimension over the test instances. 

\[ BAcc = 1 - 1/2(FPR+FNR)\]
where  $FPR=FP/(FP+TN)$ and $FNR=FN/(FN+TP)$.
FPR indicates the rate at which an evaluator incorrectly predicts that a summary sentence contains an error when it is actually correct and FNR represents the rate at which an evaluator incorrectly predicts that a summary sentence is correct when it actually contains an error.

Generally, positive shows there is a faithfulness error in the summary while negative means there is no error in the summary.   More specifically:
FP: instances where the ground truth label for the summary is 1 (faithful) but the predicted label is 0 (unfaithful)
FN: instances where the ground truth label for the summary is 0 (faithful) but the predicted label is 1 (unfaithful)

\subsubsection{Krippendorff alpha}
A good evaluator not only has to achieve high accuracy but it also has to be well-aligned with human annotations by scoring higher IAA. Hence, to measure this alignment, we use Krippendorff alpha (K-alpha) \cite{krippendorff2011computing} to measure the correlation between system and human evaluations. 

\subsection{Dataset Statistics}\label{app:data-stats}
We have used TofuEval \cite{tang2024tofueval} and AggreFact \cite{tang-etal-2023-understanding} with diaolgue and news domain instances respectively.
The statistics of the datasets we have used for our evaluations are presented in Table \ref{tab:dataset}. We report both the number of individual sentences as well as full summaries for TofuEval as we report results both on sentence-level and summary-level evaluation.
\begin{table}
\centering
\small
\begin{tabular}{@{}llccc@{}}
\toprule
Benchmark                  & Dataset     & \multicolumn{1}{l}{Sents} & \multicolumn{1}{l}{Sums} & \multicolumn{1}{l}{\% unfaithful} \\ \midrule 
\multirow{2}{*}{TofuEval}  & MediaS    & 726& 266                      & 44\%                                        \\
                           & MeetingB & 772 &266                      & 37\%                                        \\ \midrule 
\multirow{2}{*}{AggreFact} & CNN        & - & 558                      & 10\%                                        \\
                           & XSum       & - & 558                      & 49\%      \\                                 
\bottomrule
\end{tabular}
\caption{Dataset statistics with number of annotated summaries. TofuEval contains separate sentence-level annotations.}
\label{tab:dataset}
\end{table}

\section{Ambiguity}\label{app:ambiguity}
\begin{figure*}
    \centering
    \includegraphics[width=1\linewidth]{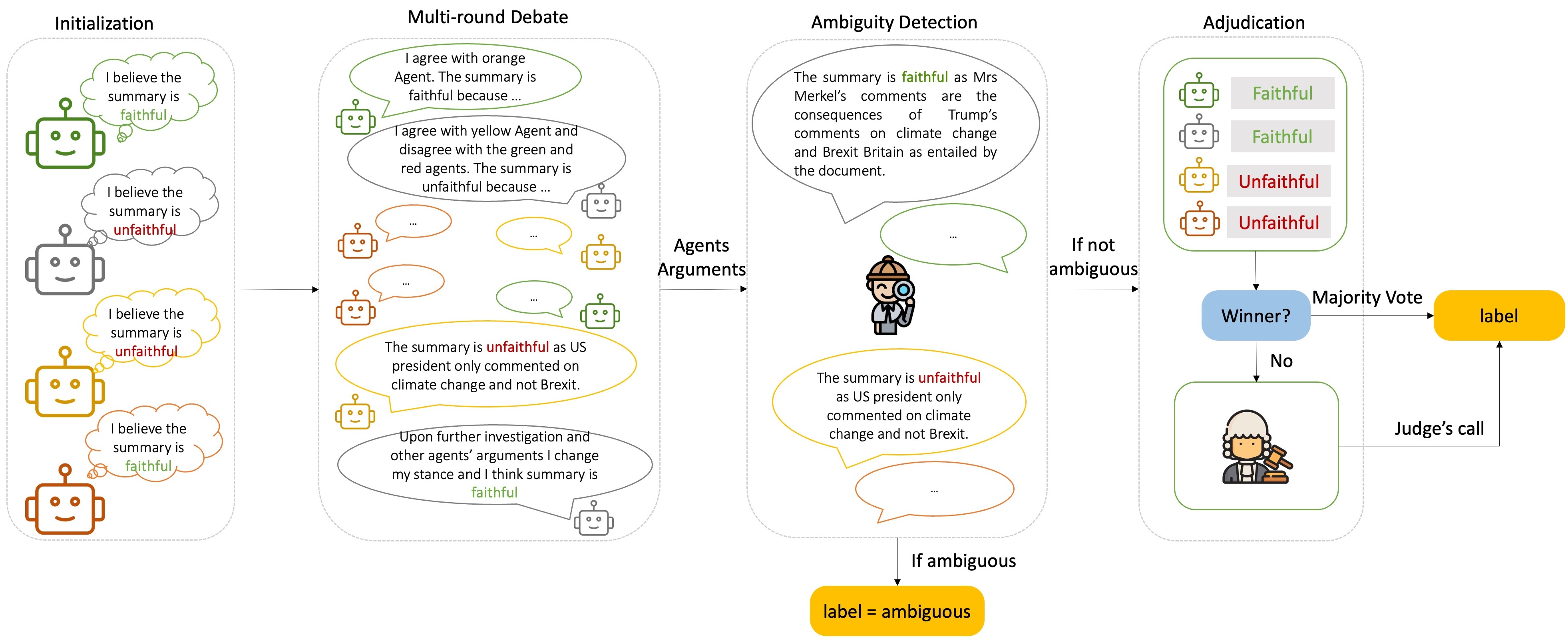}
    \caption{Faithfulness evaluator with ambiguity detection module.}
    \label{fig:overview_w_ambiguity}
\end{figure*}

An ideal faithfulness evaluation system should handle ambiguities first. This can be done by identifying the ambiguous summaries and filtering them out and then evaluating the non-ambiguous summaries. 
The overall view of a faithfulness evaluator with the ambiguity detection module is shown in Figure \ref{fig:overview_w_ambiguity}.

A summarizer can additionally be evaluated on ambiguity dimension and be provided with feedback to avoid generating ambiguous summaries. This can be seen as a future direction and is out of scope of this work.

\subsection{Ambiguity Taxonomy}\label{app:taxonomy}
We provide a detailed taxonomy of ambiguities based on our definition of ambiguity in summaries which consists of $3$ main categories and overall $16$ fine-grained sub-categories to help with ambiguity detection. 
The detailed proposed ambiguity taxonomy with definitions and examples is presented in Table \ref{tab:ambiguity2}.

\begin{table*}
\scriptsize
\centering
\begin{tabular}{@{}p{1.2cm}p{2.5cm}p{10cm}p{1cm}@{}}
\toprule
\textbf{Category}& \textbf{Type} & \textbf{Definition} & \textbf{Example}\\ 
\midrule
Implicit \newline reasoning \newline phenomena & Deduction                                                                       &   The summarizer has made a logical deduction, utilizing premises solely from the source document to draw a conclusion that cannot be directly traced to the source document. The conclusion can be accurate or inaccurate, but the key aspect of this label is that individual premises were accurately identified in the source document, and were used to form a conclusion that is stated in the summary sentence.Importantly, this deduction must be significant enough to make it noticeably harder for someone to assess the factuality of the summary sentence. Minor or obvious deductions that don't create evaluation challenges should not be included.   & Table \ref{tab:deduction_example}                                                                            \\
\cmidrule{2-4}
                             & Inference: \newline Common-sense                                                         &   The summarizer appears to have made an inference based on at least one premise from the source document, in addition to at least one premise that is based on a common-sense notion that is not stated explicitly in the source document. This common-sense notion should be widely accepted but not so universally obvious that it doesn't create any evaluation challenge. The resulting inference should make it noticeably harder to assess the factuality of the summary sentence against the source document alone.                                              & Table \ref{tab:commonsense_example}                               \\
                             \cmidrule{2-4}
                             & Inference: \newline Value-based                                                          &     The summarizer appears to have made an inference based on at least one premise from the source document in addition to at least one premise that is based on a value assumption.'Value' here specifically refers to a moral, ethical, or societal belief, principle, or ideal that guides or motivates attitudes and actions. This is distinct from common-sense notions or industry-standard evaluations. The value-based premise should be significant enough that it creates a notable challenge in evaluating the factuality of the summary sentence against the source document alone. The value assumption should not be explicitly stated in the source document but should be a recognizable societal or cultural value that the summarizer has applied to interpret the information.                                       & Table \ref{tab:value_based_example}                                           \\
                             \cmidrule{2-4}
                             & Other implicit \newline reasoning                                             &     The summarizer appears to have employed a form of implicit reasoning that goes beyond simple deduction, common-sense inference, or value-based inference, and significantly affects the summary sentence's evaluability. This category could include complex pattern recognition, synthesis of diverse information sources, experiential reasoning, or other sophisticated cognitive processes that are not explicitly traceable to the source document but are likely to have taken place in the summarizer's "mind" in order for the summary sentence to have been written. The (estimated) reasoning should be substantial enough that it creates a notable challenge in evaluating the factuality of the summary sentence against the source document alone. This category is reserved for cases that don't fit neatly into other categories of implicit reasoning but still present a clear evaluability issue due to the complexity or opacity of the reasoning process involved.                                      & Table \ref{tab:other_reasoning_example}                                            \\
\midrule
Meaning phenomena            &Semantic relations: \newline Hypernymy/Generalization                                &         A more general meaning (hypernym) is used in the summary sentence than is observed in the source document (for the same topic). This generalization should be significant enough to potentially affect the evaluability of the summary sentence.Minor generalizations that don't create evaluation challenges should not be included. The key aspect is that the generalization makes it harder to directly map the summary's claim to the specific information in the source document.   
& Table \ref{tab:hypernymy_example} \\
\cmidrule{2-4}
                             &Semantic relations: \newline Hyponymy/Specialization                                 &      A more specific meaning (hyponym) is used in the summary sentence than is observed in the source document (for the same topic). This specialization should be significant enough to potentially affect the evaluability of the summary sentence.The key aspect is that the specialization introduces details not explicitly mentioned in the source document, making it challenging to directly verify the summary's claim against the source information. Minor or widely known specializations that don't create evaluation challenges should not be included.The specialization should create a notable difficulty in assessing the factual accuracy of the summary based solely on the source document.                                     & Table  \ref{tab:hyponymy_example}                                           \\
                             \cmidrule{2-4}
                             &Semantic relations: \newline Synonymy/Paraphrasing                                   &      Meaning from the source document is paraphrased or expressed using synonyms in such a way that the summary sentence's evaluability is significantly affected. While the core meaning has not technically changed, the way the meaning is constructed or expressed has changed substantially. This paraphrasing or use of synonyms should be extensive or complex enough to create a notable challenge in directly mapping the summary's claims to the source document's information. Minor or straightforward paraphrasing that doesn't meaningfully impact evaluability should not be included. The key aspect is that the rephrasing makes it noticeably more difficult to assess the factual accuracy of the summary based solely on the source document.                                        & Table  \ref{tab:synonymy_example}                                        \\
                             \cmidrule{2-4}
                             &Linguistic ambiguity: \newline Structural                                              &     A phrase or sentence in the summary sentence has multiple valid parses (multiple valid syntactic structures), and it is not obvious which parse is intended. This ambiguity should significantly affect the evaluability of the summary sentence by creating notably different interpretations of the information presented. The key aspect is that the different possible syntactic structures lead to meaningfully different readings of the sentence, making it challenging to assess the factual accuracy of the summary against the source document. Minor ambiguities that don't substantially affect the meanings or create significant evaluation challenges should not be included, and neither should major ambiguities in which the additional interpretation is extremely implausible. The structural ambiguity should create a clear obstacle in determining which interpretation to evaluate against the source information.             & Table   \ref{tab:structural_example}                                                                  \\
                             \cmidrule{2-4}
                             &Linguistic ambiguity: \newline Lexical                                                 &     A word or phrase in the summary sentence has multiple valid interpretations in the given context, and it is not obvious which meaning is intended. This ambiguity should significantly affect the evaluability of the summary sentence by creating notably different interpretations of the information presented. The key aspect is that the different possible meanings of the word or phrase lead to meaningfully different understandings of the sentence, making it challenging to assess the factual accuracy of the summary against the source document. Minor ambiguities that don't substantially affect the overall meaning or create significant evaluation challenges should not be included. Similarly, cases where one interpretation is extremely implausible in the given context should also be excluded. The lexical ambiguity should create a clear obstacle in determining which interpretation to evaluate against the source information.      & Table  \ref{tab:lexical_example}                                                                          \\
                             \cmidrule{2-4}
                             & Vagueness                                               &         The meaning of part of the summary sentence is significantly underspecified compared to the source document, resulting in many different realities being compatible with the claim made.This vagueness should be substantial enough that:
                             \newline
1. There is confusion about what specific claim is actually being made.
\newline
2. The claim cannot be evaluated reliably against the source document.
\newline
3. The range of possible interpretations is so broad that it becomes challenging to determine if the summary accurately represents the source information.
\newline
The key aspect is that the vagueness creates a meaningful obstacle in assessing the factual accuracy of the summary.Minor instances of underspecification that don't significantly impact evaluability should not be included. The vagueness should go beyond simple generalization or summarization and create a genuine challenge in mapping the summary's claims to the specific information provided in the source document.  
& Table \ref{tab:vagueness_example}\\
                              \cmidrule{2-4}
\end{tabular}
\caption*{}
\label{tab:ambiguity}
\end{table*}

\begin{table*}
\scriptsize
\centering
\begin{tabular}{@{}p{1.2cm}p{2.5cm}p{10cm}p{1cm}@{}}
\cmidrule{2-4}
                             & Non-assertion                                &  The summary sentence does not make a clear claim or assert anything as definitively true because it is not a standard declarative sentence. Instead, it may be:
                             \newline
1. A sentence fragment or incomplete thought
\newline
2. A question (rhetorical or otherwise)
\newline
3. A plain description without any claim
\newline
4. An exclamation or interjection
\newline
5. A command or request
\newline
6. Any other type of non-declarative expression
\newline
The key aspect is that the summary does not present a statement that can be directly evaluated for factual accuracy against the source document. This creates an evaluability issue because there's no clear assertion to assess for truthfulness or correspondence with the source information, or if there is an implicit assertion, the sentence's non-declarative nature makes it hard to identify the assertion.
\newline
Note that sentences in headline style (e.g., with articles omitted) should still be considered assertions if they convey a clear, evaluable claim despite their condensed format. The focus should be on the content and function of the sentence,not just its grammatical form.    
& Table \ref{tab:non_asserstion_example}\\
                             \cmidrule{2-4}
                             & Other meaning \newline phenomenon                                &  There is something else about the literal meaning of the summary sentence that may make it challenging to assess its factuality, which is not covered by other categories in the meaning phenomena taxonomy. This could include, but is not limited to:
                             \newline
1. Use of metaphorical or highly figurative language that doesn't have a clear, literal correspondence to the source document's content.
  \newline
2. Referential ambiguities not covered by the existing ambiguity categories, such as unclear pronoun references without clear antecedents in the summary or its context.
  \newline
3. Unusual or creative uses of language that introduce interpretive challenges not captured by other categories.
  \newline
The key aspect is that these phenomena should create a significant obstacle in evaluating the factual accuracy of the summary against the source document. The issue should be substantial enough that it genuinely impedes the ability to determine if the summary accurately represents the source information.
Note that this category should only be used when the meaning phenomenon doesn't fit clearly into any other category in the taxonomy. Minor stylistic choices or common figures of speech that don't significantly impact evaluability should not be included.    
& Table \ref{tab:other_meaning_example}\\
\midrule
Context \newline phenomena            & Decontextualization                                     &                     The summary sentence presents information from the source document in a way that significantly alters its intended meaning or interpretation by removing crucial contextual elements. This can include presenting hypothetical scenarios as facts, stripping statements of important qualifications, or omitting key background information. The removal of context should create a meaningful evaluability issue by changing the meaning, losing important nuances, or making it difficult to assess the summary's factuality against the source. This issue should be substantial enough to alter the interpretation of the information, not merely a simplification that maintains the core meaning and context.                                   
& Table \ref{tab:decontextualization_example}\\
\cmidrule{2-4}
                             & Conflation                                              &    Conflation occurs when the summary sentence inappropriately combines or merges distinct pieces of information from the source document in a way that significantly affects the evaluability of the summary's factuality. This can include:
                             \newline
1. Merging separate topics or events as if they were a single issue.
 \newline
2. Combining attributes or characteristics of different entities or concepts.
 \newline
3. Blending outcomes or decisions related to distinct matters.
 \newline
The key aspect is that this merging of information creates a meaningful evaluability issue by misrepresenting relationships between different pieces of information or making it challenging to accurately assess the factuality of the combined statement. Conflation should be substantial enough to create genuine difficulty in evaluating the summary against the source document, beyond minor simplifications or simple misattributions that don't involve merging distinct concepts or information.  
& Table \ref{tab:conflation_example}\\
                             \cmidrule{2-4}
                             & Other context \newline phenomenon                                &   This category covers context-related challenges in evaluating the summary sentence's factuality that don't fit neatly into the Decontextualization or Conflation categories. These issues arise from how the summary interprets or presents the relationships between different pieces of information in the source document. Examples might include:
                             \newline
1. Inferring causal relationships not explicitly stated in the source.
\newline
2. Reordering information in a way that implies a different significance or relationship than in the original context.
\newline
3. Drawing conclusions about the overall meaning or importance of information based on its placement or context in the source document.
\newline
The key aspect is that these phenomena should create a significant obstacle in evaluating the factual accuracy of the summary against the source document, stemming from how the summary interprets or represents the context of the information. The issue should be substantial enough that it genuinely impedes the ability to determine if the summary accurately represents the source information and its intended meaning or significance.
This category should only be used when the context-related issue doesn't clearly fit into Decontextualization or Conflation,and when it creates a genuine evaluability challenge. 
& Table \ref{tab:other_context_example}
\\
\midrule
                       Other &  Other evaluability \newline issue    & This category covers evaluability challenges that don't fit into any other category in the taxonomy. These issues should significantly impede the ability to assess the factual accuracy of the summary sentence against the source document.Examples might include:
                       \newline
1. Subjective interpretations of objective information that make factual assessment difficult.
\newline
2. Reliance on specialized cultural or contextual knowledge not provided in the source document and not common enough to be considered general knowledge.
\newline
3. Novel or unique challenges in comparing the summary to the source that aren't captured by existing categories.
\newline
It's crucial to note that factual errors alone do not create an evaluability issue. The key aspect is that these phenomena should create a significant obstacle in determining whether the summary accurately represents the information in the source document.
This category should only be used when the evaluability issue doesn't clearly fit into any other category in the taxonomy and when it creates a genuine, substantial challenge in assessment.
& Table \ref{tab:other_example}
\\ \bottomrule
\end{tabular}
\caption{Ambiguity taxonomy: categories and definitions}
\label{tab:ambiguity2}
\end{table*}

\begin{table*}
\centering
\small
\begin{tabular}{@{}p{14cm}@{}}
\toprule
\textbf{\textcolor{red}{Ambiguous example: Street Closures}}
\newline
\textbf{Source document excerpt:}
The annual Bloomington Street Festival will be held on February 2nd and February 3rd this year.The event attracts thousands of visitors and features local artisans, food vendors, and live music performances.
\newline
\textbf{Summary sentence:}
Some streets will be closed on February 2-3.
\newline
\textbf{Explanation:}
This is a positive example of a Deduction evaluability issue. The summary sentence makes a claim that is not directly stated in the source document. Instead, it appears to be the result of a deduction based on two premises from the source:1) The festival is held on February 2nd and 3rd, and 2) It’s referred to as a “street festival”. While this deduction is logically valid, it creates an evaluability issue because an evaluator would likely try to verify this claim directly in the source document, where it’s not explicitly stated. The deductive leap, while reasonable, makes it harder to assess the factuality of the summary sentence against the source document. Therefore, this summary sentence should be marked with the `Deduction' label.
\\
\midrule
\textbf{\textcolor{teal}{Non-ambiguous example: Music Budget}}
\newline
\textbf{Source document excerpt:}
The Smithville School Board meeting on Tuesday addressed several topics. The board unanimously approved a budget increase of \$500,000 for the music department. This additional funding will be used to purchase new instruments, hire two part-time music teachers, and expand the after-school music program to include elementary school students.
\newline
\textbf{Summary sentence:}
The Smithville School Board has decided to invest more in music education.
\newline
\textbf{Explanation:}
This is a negative example for the Deduction evaluability issue. While the summary sentence does involve a minor deduction - connecting the approval of a budget increase for the music department with “investing more in music education” - this deduction is so straightforward and closely tied to the explicit information in the source that it doesn’t create an evaluability issue. The connection between increasing the budget for instruments, hiring teachers, and expanding programs, and “investing more in music education” is immediate and obvious. An evaluator would have no difficulty assessing the factuality of this summary sentence based on the information provided in the source document. Therefore, despite involving a slight deductive step, this summary sentence should not be marked with the ‘Deduction’ label as it doesn’t create any significant challenge in evaluation.
    \\ 
 \bottomrule
\end{tabular}
\caption{Examples for implicit reasoning phenomena: deduction. The \textcolor{red}{red example} is the one with ambiguity and the \textcolor{teal}{teal example} is a case that the deduction would not lead to ambiguities.}
\label{tab:deduction_example}
\end{table*}
\begin{table*}
\centering
\small
\begin{tabular}{@{}p{14cm}@{}}
\toprule
\textbf{\textcolor{red}{Ambiguous example: Economic Impact}}
\newline
\textbf{Source document excerpt:}
The annual Bloomington Street Festival will be held next month in the downtown area. Organizers expect over 50,000 visitors from across the state to attend the two-day event, which features local artisans, food vendors, and live music performances.
\newline
\textbf{Summary sentence:}
The festival will significantly boost local business in downtown Bloomington.
\newline
\textbf{Explanation:}
This is a positive example of a Common-sense inference evaluability issue. The summary sentence makes a claim that is not explicitly stated in the source document. Instead, it appears to be an inference based on two premises: 1) The festival will bring over 50,000 visitors to downtown Bloomington (from the source), and 2) Large events that bring many people to an area tend to increase business for local businesses (a common-sense notion). While this inference is likely valid, it creates an evaluability issue because an evaluator would need to rely on the same common-sense notion to assess its accuracy, rather than finding the information directly stated in the source. This reliance on common-sense reasoning, while often accurate, makes it harder to objectively evaluate the factuality of the summary sentence against the source document.
\\
\midrule
\textbf{\textcolor{teal}{Non-ambiguous example: Voter Influence}}
\newline
\textbf{Source document excerpt:}
Chancellor Angela Merkel delivered a speech at an election rally in Munich on Sunday, addressing key policy issues ahead of the September elections. The event was attended by thousands of supporters and local party members.
\newline
\textbf{Summary sentence:}
Merkel’s speech may influence some German voters’ opinions.
\newline
\textbf{Explanation:}
This is a negative example for the Common-sense inference evaluability issue. While the summary sentence does involve an inference based on information from the source (Merkel made a speech at an election rally) and a common-sense notion (political speeches can affect voters’ opinions), this inference doesn’t create a significant evaluability issue. The common-sense premise that political speeches can influence voters is so widely accepted and fundamental to the understanding of political rallies that it doesn’t introduce any real challenge in evaluating the summary’s factuality. The connection between giving a speech at a rally and potentially influencing voters is immediate and obvious. Therefore, despite being an inference involving common sense, this summary sentence should not be marked as having an evaluability issue, as it doesn’t create any meaningful difficulty in assessment.
    \\ 
 \bottomrule
\end{tabular}
\caption{Examples for implicit reasoning phenomena: commonsense}
\label{tab:commonsense_example}
\end{table*}
\begin{table*}
\centering
\small
\begin{tabular}{@{}p{14cm}@{}}
\toprule
\textbf{\textcolor{red}{Ambiguous example: CEO Conduct}}
\newline
\textbf{Source document excerpt:}
During his keynote speech at the industry conference, CEO John Smith of Tech Innovations Inc.referred to rival company XYZ’s new smartphone as ‘a glorified paperweight’ and mimicked throwing it into a trash can. His remarks drew gasps from some attendees, while others chuckled uncomfortably.
\newline
\textbf{Summary sentence:}
The CEO’s public ridicule of a competitor’s product at the industry conference was widely viewed as unprofessional and damaging to the company’s reputation.
\newline
\textbf{Explanation:}
This is a positive example of a Value-based inference evaluability issue. The summary sentence makes claims about how the CEO’s actions were perceived, which are not explicitly stated in the source document. This inference appears to be based on the factual information from the source (the CEO’s mocking remarks) and a value-based premise that professional conduct in business, especially for high-level executives, should involve treating competitors with respect and focusing on one’s own products rather than denigrating others’. This commonly-held value is not stated in the source but is crucial to the inference.The reliance on this value-based premise creates an evaluability issue because it requires the evaluator to share or recognize this value to assess the accuracy of the summary’s claims about how the CEO’s actions were perceived.
\\
\midrule
\textbf{\textcolor{teal}{Non-ambiguous example: Pothole Repairs}}
\newline
\textbf{Source document excerpt:}
At yesterday’s city council meeting, members unanimously approved a \$500,000 budget to fix potholes on Main Street over the next month. The decision came after numerous complaints from residents about the road’s condition.
\newline
\textbf{Summary sentence:}
The city’s decision to repair potholes on Main Street will improve road safety for drivers.
\newline
\textbf{Explanation:}
This is a negative example for the Value-based inference evaluability issue. While the summary sentence does involve an inference not explicitly stated in the source, it’s not primarily based on a value judgment. An annotator might mistakenly think this involves a value-based premise like “Safety is important and should be maintained in public spaces.” However, the inference is more accurately based on the common-sense notion that “Pothole repairs make roads safer.” This is not a moral or ethical value, but rather a widely accepted understanding of road maintenance effects. Therefore, this should be considered a common-sense inference rather than a value-based one. It doesn’t create the kind of evaluability challenge associated with value-based inferences, where personal or societal values significantly influence the interpretation of facts.
\\
\midrule
\textbf{\textcolor{teal}{Non-ambiguous example: Phone Upgrade}}
\newline
\textbf{Source document excerpt:}
The upcoming smartphone from TechCorp features a processor that is 20\% faster, a camera with 5 megapixels more resolution, and a battery that lasts 2 hours longer than the previous model. The company plans to release the new model next month.
\newline
\textbf{Summary sentence:}
The new smartphone model is expected to be a significant upgrade from its predecessor.
\newline
\textbf{Explanation:}
This is another negative example for the Value-based inference evaluability issue. An annotator might mistakenly label this as a value-based inference, thinking that “significant upgrade” involves a value judgment. However, this inference is based on commonly accepted standards in the tech industry rather than moral, ethical, or personal values. The judgment that faster processors, better cameras, and longer battery life constitute an upgrade is a technical evaluation, not a value-based one. This should be considered a common-sense inference within the context of technology advancements. The example highlights the importance of distinguishing between technical or industry-standard evaluations and true value judgments based on moral,ethical, or societal values.
    \\ 
 \bottomrule
\end{tabular}
\caption{Examples for implicit reasoning phenomena: value-based inference.}
\label{tab:value_based_example}
\end{table*}
\begin{table*}
\centering
\small
\begin{tabular}{@{}p{14cm}@{}}
\toprule
\textbf{\textcolor{red}{Ambiguous example: Investor Sentiment}}
\newline
\textbf{Source document excerpt:}
In the quarterly earnings call, CEO Jane Smith projected a 15\% revenue growth for the next fiscal year. She cited strong product performance and expansion into new markets as key drivers for this optimistic forecast. The company’s stock price saw a modest 2\% increase following the announcement.
\newline
\textbf{Summary sentence:}
Despite the CEO’s optimistic forecast, investors remain cautious about the company’s future performance.
\newline
\textbf{Explanation:}
This is a positive example of an Other implicit reasoning phenomenon. The summary sentence makes a claim about investor sentiment that isn’t explicitly stated in the source document. This inference appears to be based on a complex synthesis of information and experience that goes beyond simple deduction or common-sense reasoning. The summarizer seems to have considered factors such as past experiences with CEO projections, general market conditions, the company’s  recent performance history, and the tendency of executives to present optimistic forecasts. This type of pattern recognition and experiential reasoning creates an evaluability issue because it relies on implicit knowledge and interpretation that isn’t directly traceable to the source document. An evaluator would find it challenging to assess the factuality of the claim about investor caution based solely on the information provided in the source.
\\
\midrule
\textbf{\textcolor{teal}{Non-ambiguous example: Market Challenges}}
\newline
\textbf{Source document excerpt:}
TechCorp announced the release of its new smartphone, featuring holographic display technology.The company plans to launch the product in major markets next quarter.
\newline
\textbf{Summary sentence:}
The company’s new product launch is likely to face significant challenges in the market.
\newline
\textbf{Explanation:}
This is a negative example for the Other implicit reasoning phenomenon. While the summary sentence does involve an inference not explicitly stated in the source, it doesn’t require complex market analysis or specialized industry expertise. An annotator might mistakenly think this involves sophisticated prediction about market dynamics. However, this is actually a case of common-sense inference. The summarizer is likely basing their conclusion on the general notion that new and unusual technologies often face challenges when first introduced to the market. This is a straightforward observation based on general experience, not a complex reasoning process. Therefore, this example should be labeled as a common-sense inference rather than an Other implicit reasoning phenomenon.
\\
\midrule
\textbf{\textcolor{red}{Negative example: Recycling Reception}}
\newline
\textbf{Source document excerpt:}
The city launched a mandatory recycling program requiring residents to separate their waste into three categories: recyclables, compostables, and landfill waste. In a survey conducted by the city, 45\% of residents expressed support for the program, while 40\% opposed it, and 15\% were undecided.
\newline
\textbf{Summary sentence:}
The city’s new recycling program has been met with mixed reactions from residents.
\newline
\textbf{Explanation:}
This is a negative example for the Other implicit reasoning phenomenon. An annotator might initially think this summary involves complex analysis of public opinion or implicit knowledge of local attitudes. However, the summary sentence is actually a straightforward interpretation of explicit information provided in the source document. The survey results directly support the claim of “mixed reactions” without requiring any significant inference or implicit reasoning. This summary doesn’t create an evaluability issue because it’s directly supported by the source material. Therefore, it should not be labeled as having any evaluability issue, let alone an Other implicit reasoning phenomenon.
    \\ 
 \bottomrule
\end{tabular}
\caption{Examples for implicit reasoning phenomena: other}
\label{tab:other_reasoning_example}
\end{table*}
\begin{table*}
\centering
\small
\begin{tabular}{@{}p{14cm}@{}}
\toprule
\textbf{\textcolor{red}{Ambiguous example: Dog Popularity}}
\newline
\textbf{Source document excerpt:}
The study found that golden retrievers and labrador retrievers were the most popular dog breeds in the United States last year.
\newline
\textbf{Summary sentence:}
Retriever breeds were the most popular dogs in the U.S. in the previous year.
\newline
\textbf{Explanation:}
This is a positive example for Hypernymy/Generalization because the summary uses the more general term“retriever breeds” instead of the specific breeds mentioned in the source document. This generalization makes it slightly more challenging to evaluate the factual accuracy of the summary, as it doesn’t preserve the exact level of detail from the source.
\\
\midrule
\textbf{\textcolor{teal}{Non-ambiguous example: Emissions Policy}}
\newline
\textbf{Source document excerpt:}
The new environmental policy aims to reduce carbon emissions from factories by 30\% over the next decade.
\newline
\textbf{Summary sentence:}
The recently introduced policy targets a significant decrease in industrial pollution in the coming years.
\newline
\textbf{Explanation:}
An annotator might mistakenly identify this as a Hypernymy/Generalization issue because “industrial pollution” is a broader term than “carbon emissions from factories.” However, this is not an accurate example of the evaluability issue. The summary introduces new information (“significant decrease” and “coming years”) that isn’t directly generalizing from the source. The challenge in evaluating this summary comes more from paraphrasing and potential exaggeration rather than from using a hypernym or more general meaning.
    \\ 
 \bottomrule
\end{tabular}
\caption{Examples for meaning phenomena: Hypernymy/Generalization}
\label{tab:hypernymy_example}
\end{table*}
\begin{table*}
\centering
\small
\begin{tabular}{@{}p{14cm}@{}}
\toprule
\textbf{\textcolor{red}{Ambiguous example: Park Equipment}}
\newline
\textbf{Source document excerpt:}
The city council approved funding for new playground equipment in several local parks.
\newline
\textbf{Summary sentence:}
The council has allocated money for new swings and slides in the city’s parks.
\newline
\textbf{Explanation:}
This is a positive example of Hyponymy/Specialization because the summary sentence uses more specific terms(“swings and slides”) than the general “playground equipment” mentioned in the source document. This specialization makes it challenging to evaluate the factual accuracy of the summary, as the specific types of equipment were not mentioned in the original text.
\\
\midrule
\textbf{\textcolor{teal}{Non-ambiguous example: Apple Revenue}}
\newline
\textbf{Source document excerpt:}
Apple’s annual report showed a 15\% increase in revenue from its smartphone division.
\newline
\textbf{Summary sentence:}
Apple’s latest financial statement reveals significant growth in iPhone sales.
\newline
\textbf{Explanation:}
An annotator might consider marking this as a Hyponymy/Specialization issue because “iPhone” is indeed a more specific term than “smartphone.” However, this is not a significant enough case to cause an evaluability issue. While it’s technically a hyponym, the relationship between “iPhone” and “Apple’s smartphone” is so well-known that it doesn’t meaningfully impact the ability to evaluate the factual accuracy of the summary. The vast majority of people know that the iPhone is Apple’s only smartphone, so this specialization doesn’t introduce any real ambiguity or difficulty in assessing the statement’s accuracy.
    \\ 
 \bottomrule
\end{tabular}
\caption{Examples for meaning phenomena: Hyponymy/Specialization}
\label{tab:hyponymy_example}
\end{table*}
\begin{table*}
\centering
\small
\begin{tabular}{@{}p{14cm}@{}}
\toprule
\textbf{\textcolor{red}{Ambiguous example: Emission Law}}
\newline
\textbf{Source document excerpt:}
The new legislation mandates that all vehicles sold after 2030 must be zero-emission models.
\newline
\textbf{Summary sentence:}
The recently passed law requires that automobiles available for purchase post-2030 be free from exhaust emissions.
\newline
\textbf{Explanation:}
This is a positive example of Synonymy/Paraphrasing because the summary sentence conveys the same meaning as the source, but uses different words and sentence structure. Terms like “legislation” become “law,” “vehicles” become“automobiles,” and “zero-emission” is paraphrased as “free from exhaust emissions.” While the core meaning remains the same, the extensive paraphrasing makes it more challenging to evaluate the factual accuracy of the summary, as an evaluator would need to carefully consider whether each paraphrased element truly maintains the original meaning.
\\
\midrule
\textbf{\textcolor{teal}{Non-ambiguous example: Exercise Benefits}}
\newline
\textbf{Source document excerpt:}
The research indicates that regular exercise can significantly reduce the risk of cardiovascular disease.
\newline
\textbf{Summary sentence:}
The study shows that frequent physical activity can greatly lower the chances of heart problems.
\newline
\textbf{Explanation:}
An annotator might be tempted to mark this as a Synonymy/Paraphrasing issue because several terms have been replaced with synonyms (e.g., “exercise” with “physical activity,” “reduce” with “lower”). However, this level of paraphrasing is not significant enough to affect the evaluability of the summary sentence. The changes are straightforward and commonly understood equivalents that don’t introduce any real ambiguity or difficulty in assessing the statement’s accuracy. While paraphrasing has occurred, it doesn’t meaningfully impact the ability to evaluate the factual accuracy of the summary.
    \\ 
 \bottomrule
\end{tabular}
\caption{Examples for meaning phenomena: Synonymy/Paraphrasing}
\label{tab:synonymy_example}
\end{table*}
\begin{table*}
\centering
\small
\begin{tabular}{@{}p{14cm}@{}}
\toprule
\textbf{\textcolor{red}{Ambiguous example: University Changes}}
\newline
\textbf{Source document excerpt:}
The university announced new funding for research projects in biology and chemistry departments. Separately, they introduced stricter guidelines for laboratory safety procedures and a new online course registration system.
\newline
\textbf{Summary sentence:}
The university implemented new laboratory safety guidelines and online registration for biology courses.
\newline
\textbf{Explanation:}
This is a positive example of Structural ambiguity because the summary sentence can be parsed in two distinctly different and plausible ways: 1) The university implemented [new laboratory safety guidelines] and [online registration for biology courses]. 2) The university implemented [new laboratory safety guidelines and online registration] for [biology courses]. In the first interpretation, the university did two separate things: implemented new lab safety guidelines (potentially for all departments) and introduced online registration specifically for biology courses. In the second interpretation, both the new safety guidelines and the online registration system are specifically for biology courses. This significant difference in meaning based on the syntactic structure creates a genuine evaluability challenge, as it’s not clear which interpretation is intended and they lead to very different factual claims about the scope of the implementations.
\\
\midrule
\textbf{\textcolor{red}{Ambiguous example: Merkel’s Statement}}
\newline
\textbf{Source document excerpt:}
Mrs Merkel said she wanted friendly relations with both countries as well as Russia but Europe now had to “fight for its own destiny”. Her comments come after Mr Trump refused to re-commit to the 2015 Paris climate deal at the G7 summit. “The times in which we could completely depend on others are on the way out. I’ve experienced that in the last few days,” Mrs Merkel told a crowd at an election rally in Munich, southern Germany. The relationship between Berlin and new French President Emmanuel Macron had to be a priority, Mrs Merkel said, adding: “We Europeans have to take our destiny into our own hands.” Mr Trump has previously pledged to abandon the Paris deal, and expressed doubts about climate change.Speaking in Brussels last week, Mr Trump also told Nato members to spend more money on defence and did not re-state his administration’s commitment to Nato’s mutual security guarantees.
\newline
\textbf{Summary sentence:}
German Chancellor Angela Merkel has said the times when Europe could “completely depend on others” are “on the way out” after US President Donald Trump’s comments on climate change and Brexit.
\newline
\textbf{Explanation:}
This is a positive example of Structural ambiguity because the summary sentence can be parsed in two distinctly different ways: 1) … after [[US President Donald Trump’s comments on climate change] and [Brexit]] In this interpretation, Merkel’s statement is in response to two separate things: Trump’s comments on climate change, and Brexit. 2) … after [US President Donald Trump’s comments on [[climate change] and [Brexit]]] In this interpretation, Merkel’s statement is in response to Trump’s comments on two topics: climate change and Brexit. This ambiguity in the syntactic structure creates a significant evaluability challenge. The first interpretation suggests that Brexit itself (not Trump’s comments on it) is part of the reason for Merkel’s statement, while the second interpretation attributes both climate change and Brexit comments to Trump. These different parses lead to different factual claims about the reasons behind Merkel’s statement, making it difficult to evaluate the accuracy of the summary without clarification.
\\
\midrule
\textbf{\textcolor{teal}{Non-ambiguous example: Company Growth}}
\newline
\textbf{Source document excerpt:}
The annual report shows that the company’s profits increased by 10\% in the technology sector and 5\% in the retail sector.
\newline
\textbf{Summary sentence:}
The company saw growth in both its technology and retail divisions.
\newline
\textbf{Explanation:}
An annotator might initially consider marking this as a Structural ambiguity issue, but upon closer examination, it becomes clear that this is actually an example of Vagueness. The sentence structure itself is not ambiguous; there’s only one valid parse. However, the term “growth” is vague and underspecified compared to the precise percentages given in the source document. This vagueness allows for many realities to be compatible with the claim (any positive growth in both sectors would satisfy the summary). While this does create some evaluability challenges, it’s due to the lack of specificity rather than multiple possible syntactic structures. Therefore, this example would be better categorized under the Vagueness evaluability issue rather than Structural ambiguity.
    \\ 
\midrule
\textbf{\textcolor{teal}{Non-ambiguous example: City Achievements}}
\newline
\textbf{Source document excerpt:}
The city’s annual report highlighted two major achievements: the completion of the new public library and the successful implementation of a city-wide recycling program. Mayor Johnson praised the efforts of city employees and volunteers who contributed to these projects.
\newline
\textbf{Summary sentence:}
The mayor commended city workers and volunteers who built the library and implemented the recycling program.
\newline
\textbf{Explanation:}
An annotator might initially be tempted to mark this as a Structural ambiguity issue because the sentence could theoretically be parsed in two ways: 1) The mayor commended ((city workers and volunteers) who built the library and implemented the recycling program), or 2) The mayor (commended city workers and volunteers who built the library) and(implemented the recycling program). However, this should not be considered a case of Structural ambiguity that creates an evaluability issue. While there is technically an ambiguity in the syntactic structure, the second interpretation in which it was the mayor herself who implemented the recycling program is extremely implausible. No reasonable reader would assume this interpretation, given common knowledge about how city projects typically work. The much more likely and sensible interpretation is that it was the city workers and volunteers who built the library and implemented the recycling program. Therefore, this example doesn’t create a significant evaluability challenge and shouldn’t be labeled as having a Structural ambiguity issue.
    \\ 
 \bottomrule
\end{tabular}
\caption{Examples for meaning phenomena: Structural ambiguity}
\label{tab:structural_example}
\end{table*}
\begin{table*}
\centering
\small
\begin{tabular}{@{}p{14cm}@{}}
\toprule
\textbf{\textcolor{red}{Ambiguous example: Menu Calories}}
\newline
\textbf{Source document excerpt:}
The city’s new ordinance requires all restaurants to clearly display calorie information for each dish on their menus.
\newline
\textbf{Summary sentence:}
Local eateries must now post dishes’ calorie counts.
\newline
\textbf{Explanation:}
This is a positive example of Lexical ambiguity because the word “post” in the summary sentence has multiple valid interpretations. It could mean: 1) to physically display or affix the information in the restaurant, or 2) to publish or upload the information online. Without additional context, it’s not obvious which meaning is intended. This ambiguity makes it challenging to evaluate the factual accuracy of the summary, as the source document specifically mentions displaying the information on menus, but the summary’s use of “post” leaves room for different interpretations.
\\
\midrule
\textbf{\textcolor{teal}{Non-ambiguous example: Ancient Knowledge}}
\newline
\textbf{Source document excerpt:}
The latest archaeological findings suggest that the ancient civilization had advanced knowledge of astronomy and mathematics.
\newline
\textbf{Summary sentence:}
Recent discoveries indicate the early society was versed in celestial and numerical sciences.
\newline
\textbf{Explanation:}
An annotator might be tempted to mark this as a Lexical ambiguity issue because of the use of less common terms like “versed” or “celestial sciences”. However, this is not a true case of lexical ambiguity. While the words used in the summary are more formal or academic, they don’t have multiple valid interpretations in this context. “Versed” clearly means knowledgeable or skilled, and “celestial sciences” is an obvious reference to astronomy. The words, though perhaps less common, have singular,clear meanings in this context. Any perceived difficulty in evaluation comes from the use of synonyms or domain-specific language, not from words having multiple possible interpretations. Therefore, this example doesn’t present a lexical ambiguity evaluability issue.
    \\ 
 \bottomrule
\end{tabular}
\caption{Examples for meaning phenomena: Lexical ambiguity}
\label{tab:lexical_example}
\end{table*}
\begin{table*}
\centering
\small
\begin{tabular}{@{}p{14cm}@{}}
\toprule
\textbf{\textcolor{red}{Ambiguous example: Exercise Effects}}
\newline
\textbf{Source document excerpt:}
A longitudinal study tracking 10,000 adults over 20 years found that those who engaged in regular physical activity (defined as at least 150 minutes of moderate exercise per week) had a 37\% lower risk of developing type 2 diabetes compared to sedentary individuals. The study controlled for factors such as diet, family history, and initial BMI.
\newline
\textbf{Summary sentence:}
Research suggests that being active may have health benefits.
\newline
\textbf{Explanation:}
This is a positive example of vagueness because the summary sentence is extremely underspecified compared to the source. The terms “being active,” “may have,” and “health benefits” are so vague that they’re compatible with an enormous range of realities. What counts as “being active”? How probable is “may”? What specific “health benefits” are we talking about? The vagueness is so significant that it’s unclear what specific claim is being made, making it very difficult to evaluate the accuracy of the summary against the precise information in the source document. This level of vagueness creates a genuine evaluability issue, as the summary could be considered technically true for even minimal activity providing any small health benefit, but the source document hones in on much more specific health benefits for specific conditions.
\\
\midrule
\textbf{\textcolor{teal}{Non-ambiguous example: EV Performance}}
\newline
\textbf{Source document excerpt:}
The new electric vehicle model has a range of 400 miles on a single charge, compared to the current industry average of 250 miles. It can accelerate from 0 to 60 mph in 3.5 seconds, while the average for electric cars in its class is 6 seconds. These specifications place it in the top 1\% of electric vehicles currently on the market in terms of both range and acceleration.
\newline
\textbf{Summary sentence:}
The latest electric car boasts impressive range and acceleration capabilities.
\newline
\textbf{Explanation:}
An annotator might initially consider marking this as a vagueness issue because terms like “impressive” and“capabilities” are not as specific as the numbers in the source document. However, upon examination of the source, it’s clear that this level of vagueness doesn’t create a significant evaluability issue. The use of “impressive” is justified by the explicit comparisons to industry averages and the car’s placement in the top 1\% for both metrics. While the summary doesn’t provide exact figures, it doesn’t create confusion about what claim is being made or make evaluation unreliable. The characterization of the range and acceleration as “impressive” is fair and evaluable against the specific comparisons provided in the source document. This example demonstrates acceptable summarization rather than problematic vagueness.
    \\ 
 \bottomrule
\end{tabular}
\caption{Examples for meaning phenomena: Vagueness}
\label{tab:vagueness_example}
\end{table*}
\begin{table*}
\centering
\small
\begin{tabular}{@{}p{14cm}@{}}
\toprule
\textbf{\textcolor{red}{Ambiguous example: Zoning Question}}
\newline
\textbf{Source document excerpt:}
The city council voted 7-2 in favor of the new zoning ordinance, which will allow for the construction of multi-family housing units in previously single-family residential areas. Proponents argue this will help address the city’s housing shortage, while opponents express concerns about increased traffic and changes to neighborhood character.
\newline
\textbf{Summary sentence:}
Will the new zoning law really solve the housing crisis?
\newline
\textbf{Explanation:}
This is a positive example of Non-assertion because the summary sentence is a question rather than a declarative statement. It doesn’t assert any facts or make any claims about the zoning law or its effects. Instead, it poses a rhetorical question that cannot be evaluated for factual accuracy against the source document. This creates an evaluability issue because there’s no clear assertion to assess – the question merely raises a point for consideration without providing any information that can be judged as true or false.
\\
\midrule
\textbf{\textcolor{red}{Ambiguous example: Economic Data}}
\newline
\textbf{Source document excerpt:}
The annual economic report released by the Federal Reserve indicates that inflation rates have decreased from 6.5\% to 3.2\% over the past year, while unemployment has remained stable at 3.8\%. The report suggests that these trends reflect a gradual stabilization of the economy following recent global disruptions.
\newline
\textbf{Summary sentence:}
Promising economic indicators
\newline
\textbf{Explanation:}
This is a positive example of Non-assertion because the summary sentence is a mere phrase rather than a complete sentence. It doesn’t make any explicit claims or assertions about the economic situation. While it implies that there are economic indicators and that they are promising, it doesn’t actually state this as a fact. The phrase could be a title, a category label, or a fragment of a longer thought. Without a verb or a complete sentence structure, there’s no clear assertion that can be evaluated for factual accuracy against the source document. This creates an evaluability issue because there’s no specific claim being made – the phrase merely suggests a topic without providing any information that can be judged as true or false.
\\
\midrule
\textbf{\textcolor{teal}{Non-ambiguous example: Whale Recovery}}
\newline
\textbf{Source document excerpt:}
A recent study by marine biologists has found that the population of blue whales in the Eastern Pacific has increased by 20\% over the past decade, attributed to strict international whaling bans and protected marine areas.
\newline
\textbf{Summary sentence:}
Scientists Report Encouraging Trend in Blue Whale Numbers
\newline
\textbf{Explanation:}
An annotator might be tempted to mark this as a Non-assertion issue because it’s written in headline style, lacking the article “an” before “encouraging trend.” However, despite its condensed form, this sentence still functions as a declarative statement with clear assertive force. It makes a specific claim that can be evaluated against the source document: scientists have reported a trend, and this trend is encouraging for blue whale populations. The omission of the article is a common feature in headlines, and while this might not be a typical style for a summary, its style doesn’t negate the sentence’s declarative nature.The increase in whale population described in the source can reasonably be characterized as “encouraging.” Therefore, this is not a case of Non-assertion, as it does make a claim that can be evaluated for accuracy, despite its headline-style formatting.
    \\ 
 \bottomrule
\end{tabular}
\caption{Examples for meaning phenomena: non-assertion}
\label{tab:non_asserstion_example}
\end{table*}
\begin{table*}
\centering
\small
\begin{tabular}{@{}p{14cm}@{}}
\toprule
\textbf{\textcolor{red}{Ambiguous example: Economic Slowdown}}
\newline
\textbf{Source document excerpt:}
The latest economic report shows that the country's GDP growth has slowed to 1.2\% in the last quarter, down from 2.8\% in the previous quarter. Analysts attribute this decrease to global supply chain disruptions and increasing energy costs. The central bank has indicated it may consider adjusting interest rates in response to these trends.
\newline
\textbf{Summary sentence:}
The economy is navigating choppy waters as growth figures take a dive.
\newline
\textbf{Explanation:}
This is a positive example of an "other meaning phenomenon" because it uses metaphorical language that creates an evaluability issue. The phrases "navigating choppy waters" and "take a dive" are figurative expressions that, while evocative,don't have a clear, literal correspondence to the economic data presented in the source document. This use of metaphor makes it challenging to assess the factuality of the summary sentence against the precise figures and factual statements in the source.While the metaphors generally align with the idea of economic difficulty and declining growth, they introduce a level of interpretive ambiguity that goes beyond the categories we've discussed. It's not a case of generalization, specialization, paraphrasing,ambiguity, vagueness, or non-assertion, but rather a use of figurative language that affects the ability to directly evaluate the claim against the source material.
\\
\midrule
\textbf{\textcolor{red}{Ambiguous example: Surprising Endorsement}}
\newline
\textbf{Source document excerpt:}
In a surprising turn of events, Senator Jane Smith publicly endorsed her long-time rival, Governor Tom Brown, for the upcoming presidential election. This endorsement comes just weeks after Brown’s campaign criticized Smith’s voting record on healthcare reform. Political analysts suggest this move could significantly impact voter perceptions in key swing states.
\newline
\textbf{Summary sentence:}
The two politicians have a history of disagreement, so her endorsement of him shocked many.
\newline
\textbf{Explanation:}
This is a positive example of an “other meaning phenomenon,” specifically a kind of referential ambiguity, which is not included in the list of ambiguity types. The summary sentence uses pronouns “her” and “him” without clear antecedents within the sentence itself or the preceding context of the summary. While readers familiar with the source document could figure out that“her” refers to Senator Smith and “him” to Governor Brown, this is not explicit in the summary sentence or preceding context.This ambiguity in pronoun reference creates an evaluability issue because it’s not immediately clear who is endorsing whom,making it challenging to assess the factual accuracy of the statement against the source document. This type of referential ambiguity doesn’t fit neatly into the previously discussed categories but significantly affects the ability to evaluate the summary’s factuality, thus qualifying as an “other meaning phenomenon.”
\\
\midrule
\textbf{\textcolor{teal}{Non-ambiguous example: Whale Population}}
\newline
\textbf{Source document excerpt:}
A recent study conducted by marine biologists at the Pacific Oceanic Institute has revealed that the population of blue whales in the Eastern Pacific Ocean has increased by approximately 20\% over the past decade. Researchers attribute this growth to the effectiveness of international whaling bans and the establishment of protected marine areas. Dr. Sarah Johnson, lead author of the study, stated, “This is a promising sign for the species’ recovery, but continued conservation efforts are crucial.”
\newline
\textbf{Summary sentence:}
Blue whale numbers in the Eastern Pacific have shown a significant uptick, according to new research.
\newline
\textbf{Explanation:}
An annotator might be tempted to categorize this as “Other meaning phenomenon” due to the use of the colloquial term “uptick” to describe the population increase. They might argue that this informal term creates an evaluability issue because it’s not as precise as the percentage given in the source document. However, this is not a valid case for the “Other meaning phenomenon” category, nor does it present a significant evaluability issue. While the use of “uptick” is an example of paraphrasing, it is not so substantial as to warrant categorization under the Synonymy/Paraphrasing evaluability issue. The term“uptick” clearly conveys the idea of increase, and “significant” accurately reflects the 20\% growth mentioned in the source. The summary sentence makes a clear, evaluable claim that aligns with the information provided in the source document without introducing any meaningful ambiguity or difficulty in assessment. Therefore, this example does not qualify as an “other meaning phenomenon” and should not be marked as having any evaluability issue.
    \\ 
\midrule
\textbf{\textcolor{teal}{Non-ambiguous example: Job Market Trends}}
\newline
\textbf{Source document excerpt:}
A recent economic report shows that the unemployment rate has dropped from 5.2\% to 4.8\% over the past six months. During the same period, the number of job openings increased by 15\%, with the technology and healthcare sectors showing the strongest growth. However, wage growth remained stagnant at 1.5\% annually, barely keeping pace with inflation.
\newline
\textbf{Summary sentence:}
The job market is improving, but workers aren’t feeling the benefits yet.
\newline
\textbf{Explanation:}
An annotator might be tempted to categorize this as an “other meaning phenomenon” due to the somewhat abstract nature of “feeling the benefits.” They might argue that this creates an evaluability issue because it’s not a direct representation of the data in the source document. However, this is not a case of an “other meaning phenomenon.” Instead, this summary sentence contains a clear example of a common-sense inference. The first part of the sentence, “The job market is improving,” is a reasonable inference based on the lower unemployment rate and increased job openings mentioned in the source. The second part, “workers aren’t feeling the benefits yet,” is an inference based on the stagnant wage growth information. These inferences rely on common-sense connections between economic indicators and their real-world impacts.Therefore, this example should be labeled as a common-sense inference under the implicit reasoning category, rather than an“other meaning phenomenon” or any other category in the meaning phenomena taxonomy.
    \\ 
 \bottomrule
\end{tabular}
\caption{Examples for meaning phenomena: Other meaning phenomenon}
\label{tab:other_meaning_example}
\end{table*}
\begin{table*}
\centering
\small
\begin{tabular}{@{}p{14cm}@{}}
\toprule
\textbf{\textcolor{red}{Ambiguous example: Cyberattack Warning}}
\newline
\textbf{Source document excerpt:}
In a hypothetical scenario presented during a cybersecurity conference, Dr. Jane Smith, a leading expert in the field, stated, “If a nation-state were to launch a coordinated cyberattack on our power grid, it could potentially leave millions without electricity for weeks.” She emphasized that this was a worst-case scenario used to illustrate the importance of robust cybersecurity measures, not a prediction of an imminent threat.
\newline
\textbf{Summary sentence:}
An expert warned that millions could be left without power for weeks due to cyberattacks.
\newline
\textbf{Explanation:}
This is a positive example of Decontextualization because the summary sentence presents Dr. Smith's statement outside of its crucial context. In the source, it’s clear that she was discussing a hypothetical scenario in a specific setting (a cybersecurity conference) to illustrate a point. The summary, however, presents it as a general warning about a real threat,stripping away the hypothetical nature and the purpose of the example. This decontextualization significantly changes the meaning and urgency of the statement, making it challenging to evaluate the factuality of the summary against the source document. The summary takes on a new, more alarming meaning when removed from its original context, creating a clear evaluability issue.
\\
\midrule
\textbf{\textcolor{teal}{Non-ambiguous example: Coffee Memory Study}}
\newline
\textbf{Source document excerpt:}
new study published in the Journal of Nutrition examined the effects of coffee consumption on cognitive function in adults over 65. The researchers found that participants who drank 2-3 cups of coffee daily showed a 15\%improvement in short-term memory tests compared to non-coffee drinkers. However, the study’s authors cautioned that more research is needed to establish a causal relationship and to account for other lifestyle factors that might influence cognitive health.
\newline
\textbf{Summary sentence:}
Recent research suggests moderate coffee consumption may boost short-term memory in older adults.
\newline
\textbf{Explanation:}
An annotator might be tempted to label this as Decontextualization because the summary doesn't mention the study’s limitations or the need for further research. However, this is not a true case of Decontextualization. The summary sentence accurately represents the main finding of the study within its proper context. It uses the word “suggests” to indicate that the relationship is not definitively proven, which aligns with the cautionary note in the source. The summary doesn't remove the information from its research context or change its meaning. While it doesn't include all the details from the source, it presents a fair and contextualized summary of the key finding. Therefore, this example doesn't create an evaluability issue due to Decontextualization and shouldn't’t be labeled as such.
    \\ 
 \bottomrule
\end{tabular}
\caption{Examples for context phenomena: Decontextualization}
\label{tab:decontextualization_example}
\end{table*}
\begin{table*}
\centering
\small
\begin{tabular}{@{}p{14cm}@{}}
\toprule
\textbf{\textcolor{red}{Ambiguous example: Library Funding}}
\newline
\textbf{Source document excerpt:}
A recent city council meeting addressed two separate issues. First, the council discussed plans to increase funding for public libraries by 10\% in the next fiscal year. In the next agenda item, they debated a proposal to extend park hours during summer months. The library funding increase was approved unanimously, while the park hours extension was tabled for further discussion due to concerns about increased maintenance costs.
\newline
\textbf{Summary sentence:}
The city council approved a measure to enhance public library access, extending their hours.
\newline
\textbf{Explanation:}
This is a positive example of Conflation because the summary sentence incorrectly combines two separate pieces of information from the source document. While the council did approve increased funding for libraries, there was no mention of extending library hours. The idea of extending hours was actually related to parks, not libraries, and that proposal was tabled, not approved. By conflating the approved library funding with the unapproved (and unrelated) extension of park hours, and then misapplying this to libraries, the summary creates a statement that is difficult to evaluate against the source document. This conflation of distinct issues and their outcomes results in a summary that misrepresents the council’s actions, making it challenging to assess its factuality. The synthesis of these separate pieces of information, along with the misattribution of the hours extension, creates a clear evaluability issue.
\\
\midrule
\textbf{\textcolor{teal}{Non-ambiguous example: Climate Change Effects}}
\newline
\textbf{Source document excerpt:}
A new study on climate change impacts has found that average global temperatures have risen by 1.1°C since pre-industrial times. The same research indicates that sea levels have risen by an average of 8 inches over the past century. The study’s authors emphasize that these two phenomena are interconnected, with rising temperatures contributing to thermal expansion of the oceans and melting of land-based ice.
\newline
\textbf{Summary sentence:}
Research shows that global warming has led to both higher temperatures and rising sea levels.
\newline
\textbf{Explanation:}
An annotator might be tempted to label this as Conflation because the summary sentence combines information about temperature rise and sea level increase into a single statement. However, this is not a true case of Conflation that creates an evaluability issue. The summary accurately represents the relationship between these phenomena as presented in the source document. The source explicitly states that these issues are interconnected, and the summary maintains this context. The synthesis of this information in the summary doesn't make it harder to evaluate its factuality against the source; rather, it provides a concise and accurate representation of the key findings and their relationship. Therefore, this example doesn't create an evaluability issue due to Conflation and shouldn't be labeled as such.
    \\ 
    \midrule
\textbf{\textcolor{teal}{Non-ambiguous example: Emissions Statement}}
\newline
\textbf{Source document excerpt:}
Speaker 0: Good evening, everyone. Today we’re joined by Dr. Emily Chen, a climate scientist, and Mr. John Davis, an energy policy expert. Speaker 1: Thank you for having me. Recent data shows that global carbon emissions have increased by 2\% in the past year, despite international efforts to reduce them. Speaker 2: That’s concerning. From a policy perspective, we need to incentivize a faster transition to renewable energy sources. Speaker 1: I agree. Our models predict that if this trend continues, we could see a 3°C rise in global temperatures by 2100. Speaker 2: That would have devastating consequences. We need to act now to prevent this.
\newline
\textbf{Summary sentence:}
John Davis stated that recent data shows a 2\% increase in global carbon emissions over the past year.
\newline
\textbf{Explanation:}
An annotator might be tempted to label this as Conflation because the summary sentence attributes a statement tot he wrong speaker, seemingly combining information from different parts of the conversation. However, this is not a true case of Conflation that creates an evaluability issue. While the summary incorrectly attributes Dr. Chen’s statement about carbon emissions to Mr. Davis, this misattribution doesn't make the sentence inherently harder to evaluate for factuality. An evaluator can easily check the source document to see who actually made the statement about the 2\% increase in emissions. The content of the statement itself is accurately represented; it’s only the speaker attribution that’s incorrect. This type of error is more akin to a simple factual mistake rather than a conflation of information that creates an evaluability issue.
Moreover, it’s important to note that the Conflation label should be reserved for cases in which ideas, concepts, topics, or other substantive content are merged or synthesized in a way that creates evaluability issues. It should not be applied to lower-level misattributions such as incorrect speaker identification or other similar factual errors. In this case, the core information and ideas presented by the speakers remain distinct and are not conflated; only the attribution of who said what is incorrect. Therefore, this example shouldn't’t be labeled as Conflation or as having any other evaluability issue.
    \\
 \bottomrule
\end{tabular}
\caption{Examples for context phenomena: Conflation}
\label{tab:conflation_example}
\end{table*}
\begin{table*}
\centering
\small
\begin{tabular}{@{}p{14cm}@{}}
\toprule
\textbf{\textcolor{red}{Ambiguous example: Parking Solution}}
\newline
\textbf{Source document excerpt:}
The city’s annual report included several sections. In the “Challenges” section, it stated: “Downtown parking remains a significant issue, with demand often exceeding supply during peak hours.” Later, in the “Future Plans” section,it mentioned: “A new multi-story parking garage is scheduled to begin construction next year, which will add 500 parking spaces to the downtown area.”
\newline
\textbf{Summary sentence:}
The city is building a new parking garage downtown because parking demand exceeds supply during peak hours.
\newline
\textbf{Explanation:}
This is a positive example of an “Other context phenomenon” because the summary sentence creates a causal relationship between two pieces of information that were presented in separate contexts within the source document. While the parking issue and the new garage construction are both mentioned, the source does not explicitly state that one is the direct cause of the other. The summary’s assertion of causality (“because”) goes beyond what’s stated in the source and creates an evaluability issue. This doesn't fit neatly into Decontextualization or Conflation categories. Instead, it represents a different kind of context-related challenge where the summarizer has inferred a relationship between separate pieces of information based on their appearance in the same document, even though they were presented in different sections with different purposes (one describing challenges, the other outlining future plans).
\\
\midrule
\textbf{\textcolor{teal}{Non-ambiguous example: Company Outlook}}
\newline
\textbf{Source document excerpt:}
In a press conference, the CEO stated: “Our company’s profits have increased by 15\% this quarter.”Later, in response to a journalist’s question about potential layoffs, she said: “We have no plans for layoffs at this time. In fact,we’re looking to expand our workforce in the coming months.”
\newline
\textbf{Summary sentence:}
Despite increased profits, the company has no plans for layoffs and is looking to hire more employees.
\newline
\textbf{Explanation:}
An annotator might be tempted to label this as an “Other context phenomenon” because the summary sentence brings together information from two different parts of the press conference and seems to imply a contrast (“Despite”). However,this is not a true case of a context-related evaluability issue. The summary accurately represents the information provided in the source document without changing its meaning or creating any challenges in evaluation. The use of “Despite” to connect the two pieces of information is a reasonable interpretation that doesn't misrepresent the context or create any evaluability issues. The relationship between the profit increase and the hiring plans is implied but doesn't distort the original information or its context.Therefore, this example shouldn't be labeled as having any context-related evaluability issue.
    \\ 
 \bottomrule
\end{tabular}
\caption{Examples for context phenomena: Other context phenomenon}
\label{tab:other_context_example}
\end{table*}
\begin{table*}
\centering
\small
\begin{tabular}{@{}p{14cm}@{}}
\toprule
\textbf{\textcolor{red}{Ambiguous example: Zoning Vote}}
\newline
\textbf{Source document excerpt:}
The city council voted 7-2 in favor of the new zoning ordinance. Council member Johnson, who had previously expressed reservations, was absent due to illness.
\newline
\textbf{Summary sentence:}
The controversial zoning ordinance passed despite opposition.
\newline
\textbf{Explanation:}
This example creates an evaluability issue while also containing factual inaccuracies. The summary’s claim that the ordinance was “controversial” is not explicitly stated in the source and should be judged as factually inaccurate (separately from the evaluability issue). The phrase “despite opposition” creates an evaluability issue because while the 7-2 vote and mention of a councilmember’s previous reservations imply some level of opposition, the extent and nature of this opposition are not clearly defined in the source. This makes it challenging to evaluate the accuracy of the “despite opposition” claim. The summary thus combines a factual inaccuracy with an evaluability issue, demonstrating how implied information can complicate the assessment of a summary’s factual accuracy.
\\
\midrule
\textbf{\textcolor{red}{Ambiguous example: Art Sale}}
\newline
\textbf{Source document excerpt:}
The painting, a large canvas covered entirely in red paint, sold for \$3 million at auction.
\newline
\textbf{Summary sentence:}
A simplistic artwork fetched an outrageously high price at the recent sale.
\newline
\textbf{Explanation:}
This example creates an evaluability issue because it interprets objective information (the description and price of the painting) in a subjective manner. The terms “simplistic” and “outrageously high” are subjective judgments not present in the source, and while it’s probably true that most people would agree with these subjective statements, their presence in the summary sentence makes it difficult to evaluate the factual accuracy of the summary sentence.
\\
\midrule
\textbf{\textcolor{red}{Ambiguous example: Festival Timing}}
\newline
\textbf{Source document excerpt:}
The local council of Whittlesea, Victoria, has announced that this year’s Whittlesea Country Music Festival will take place on the second weekend of February, as is tradition.
\newline
\textbf{Summary sentence:}
Whittlesea’s annual music event is scheduled for the height of Australian summer, potentially impacting attendance.
\newline
\textbf{Explanation:}
This example creates an evaluability issue because it requires cultural and geographical knowledge not provided in the source document. The summary makes claims about the Australian summer season and its potential impact on the event,which aren't mentioned in the source. To evaluate the accuracy of this summary, one would need to know:
\newline
1. That February is indeed summer in Australia (opposite to the Northern Hemisphere).
\newline
2. That summer in Victoria can be extremely hot, potentially affecting outdoor events.
\newline
3. The typical weather patterns in Whittlesea specifically.
This information isn't common knowledge for many people outside Australia, and it’s not provided in the source. An evaluator would need to do external research to verify these claims, making it challenging to assess the summary’s factual accuracy based solely on the given source document. This goes beyond simple paraphrasing or inference, creating a unique evaluability issue related to cultural and contextual knowledge.
\\
\midrule
\textbf{\textcolor{teal}{Non-ambiguous example: Brain Chemistry}}
\newline
\textbf{Source document excerpt:}
The study found a significant increase in dopamine levels in the nucleus accumbens following the experimental treatment.
\newline
\textbf{Summary sentence:}
The research showed the treatment boosted feel-good neurotransmitters in the brain’s reward center.
\newline
\textbf{Explanation:}
An annotator might be tempted to label this as an “Other evaluability issue” due to the use of lay terms to describe technical concepts. They might argue that this creates a unique challenge in evaluating the summary’s accuracy. However, this is actually a clear case of Synonymy/Paraphrasing. The summary translates technical terms (“dopamine” and “nucleus accumbens”) into more accessible language (“feel-good neurotransmitters” and “brain’s reward center”). While this does require some specialized knowledge to evaluate, it doesn't create a new type of evaluability issue. Instead, it falls squarely within the existing category of Synonymy/Paraphrasing, where the challenge is in determining if the paraphrased terms accurately represent the original technical language.
    \\ 
\midrule
\textbf{\textcolor{teal}{Non-ambiguous example: Product Availability}}
\newline
\textbf{Source document excerpt:}
The company announced its new product line on March 15, 2023. CEO Jane Smith stated, “We expect to begin shipping these products to customers within six months.”
\newline
\textbf{Summary sentence:}
The company’s new products are now available to customers.
\newline
\textbf{Explanation:}
An annotator might be inclined to classify this as an “Other evaluability issue” due to the temporal ambiguity created by the word “now” in the summary. They might argue that this creates a unique challenge in evaluation because the accuracy of the statement depends on when the summary was written or read. However, this is actually an example of an “Other meaning phenomenon” rather than a distinct evaluability issue. The challenge here stems from the context-dependent meaning of “now,” which is a semantic issue related to deixis (words whose meaning depends on the context of utterance). This fits within the existing category of meaning-related phenomena and doesn't constitute a new type of evaluability issue outside the current taxonomy.
    \\ 
 \bottomrule
\end{tabular}
\caption{Examples for other ambiguities: Other evaluability issue}
\label{tab:other_example}
\end{table*}

\subsection{Data Annotation for Ambiguity}\label{app:ambiguity-data-stats}
We used an existing faithfulness dataset, TofuEval MeetingBank \cite{tang2024tofueval} and annotated the sentence summaries for ambiguity.
The instructions provided to the expert annotators are presented in Table \ref{tab:annotator_intsructs}.
\begin{table*}
\centering
\small
\begin{tabular}{@{}p{14cm}@{}}
\toprule
\textbf{Overview:}
\newline
The goal of this task is to gather information about the factors or characteristics in a set of summaries (analyzed on a sentence-by-sentence basis) that impact the summary’s evaluability — the ease with which its factuality can be assessed while utilizing the summary’s source document as the source of truth. There is a diverse set of characteristics that might impact a summary’s evaluability, and we have done our best to capture a wide range of these characteristics in a Taxonomy of evaluability issues.These have been built mostly from the ground up (based on real examples), so they may not cover all kinds of evaluability issues.
\newline
\newline
This task will ultimately support the development of an automatic LLM-based summary evaluation pipeline. LLM-based factuality evaluation still lags behind humans, and one of the kinds of cases LLMs struggle with are summaries (or summary sentences)with low evaluability.
\newline
\newline

\textbf{Evaluability vs Factuality}
\newline
These two concepts are related but fundamentally distinct, and for this task, it is essential that you grasp the difference between them.
\newline
\newline
\textbf{Factuality:}
\newline
The truthfulness of a statement (or a union of statements) relative to a source of truth.
\newline
\newline
\textbf{Evaluability:}
How readily a factual evaluator can assess the factuality of the statement relative to the designated source of truth.
\newline
\newline
Along these lines, there are three pre-requisites to assessing factuality in this way:
\newline
\newline
1. A sufficiently sophisticated evaluator, with the ability for (2) and (3)
\newline
2. Comprehension of the source of truth
\newline
3. Comprehension of the statement
\newline
\newline
In principle, then, there are three variables that could prevent successful factual evaluation:
\newline
\newline
1. The evaluator is not sophisticated enough for complete comprehension/evaluation
\newline
2. The source of truth poses barriers to comprehension/evaluation
\newline
3. The statement poses barriers to comprehension/evaluation
\newline
\newline
These three variables are inter-related. For instance, (1) depends on the severity of the barriers noted in (2) and (3). But for the current task, the primary focus will be on (3), although you will also be registering your impressions for (2), and to a lesser extent,(1). (Which of the above three items is the variable that currently prevents us from using LLMs to check factuality?)
\newline
\newline
\textbf{“The statement poses barriers to comprehension”}
\newline
\newline
As mentioned above, the main focus of the current task is (3) above, so we need a way to talk about and categorize these barriers. We will use the term evaluability issue to distinguish the kinds of barriers described by (3) from the kinds of barriers described by all of (1-3). The evaluability issues we've devised fall under three main categories, plus an overflow category:
\newline
\newline
1. Implicit reasoning phenomena
\newline
2. Meaning phenomena
\newline
3. Context phenomena
\newline
4. (Other phenomena)
\newline
\newline
Refer to all the subtypes of these categories in the table below (taxonomy table as shown in Table \ref{tab:ambiguity2}). Carefully read all the definitions and refer to the examples provided in the last column. When referring to examples, ambiguous examples are examples that are an instance of the subtype, and non-ambiguous examples are examples that are not
a good example of the subtype.
    \\ 
 \bottomrule
\end{tabular}
\caption{Instructions provided to the expert annotators for ambiguity annotation.}
\label{tab:annotator_intsructs}
\end{table*}

The fine-grained data statistics is shown in Table \ref{tab:ambiguity_stats_fine}. The final dataset has a high inter-annotator agreement (Cohen's Kappa) of $\approx0.73$ on binary labels.  
The initial stage (before the adjudication step) has an IAA of $\approx0.40$ which highlights the importance of the adjudication step to achieve high-quality data.

\begin{table*}
\small
\centering
\begin{tabular}{@{}l|l|c@{}}
\toprule
\textbf{Category}& \textbf{Type} & \textbf{Number of instances} \\ 
\midrule
Implicit reasoning phenomena & Deduction &  29                                           \\
\cmidrule{2-3}
& Inference: Common-sense  &     30        \\
                             \cmidrule{2-3}
                             & Inference:  Value-based& 1\\
                             \cmidrule{2-3}
                             & Other implicit reasoning& 0\\
\midrule
Meaning phenomena            &Semantic relations:  Hypernymy/Generalization                                & 1\\
\cmidrule{2-3}
&Semantic relations: Hyponymy/Specialization& 0                                               \\
                             \cmidrule{2-3}
                             &Semantic relations: Synonymy/Paraphrasing                                   & 5    \\
                             \cmidrule{2-3}
                             &Linguistic ambiguity: Structural                                              &   5 \\
                             \cmidrule{2-3}
                             &Linguistic ambiguity: Lexical                                                 &   11                                    \\
                             \cmidrule{2-3}
                             & Vagueness&  8\\
\cmidrule{2-3}
                             & Non-assertion                                & 2\\
                             \cmidrule{2-3}
                             & Other meaning phenomenon                                & 8 \\
\midrule
Context phenomena            & Decontextualization                                     &       14              \\
\cmidrule{2-3}
                             & Conflation                                              &  28 \\
                             \cmidrule{2-3}
                             & Other context phenomenon                                & 3
\\
\midrule
                       Other &  Other evaluability issue    & 10
\\ \bottomrule
\end{tabular}
\caption{Ambiguity annotated dataset fine-grained statistics. For each sub-category, the number of such instances in the dataset is shown.}
\label{tab:ambiguity_stats_fine}
\end{table*}

\section{Prompts}\label{app:prompts}
We listed all the prompts we have used for our experiments in Table \ref{tab:prompt_lists}.
\begin{table}
\centering
\begin{tabular}{@{}l|l@{}}
\toprule
\textbf{Prompt}  & \textbf{Reference}
\\ 
\midrule
Zero-shot baseline & Table \ref{tab:direc_prompt}
\\
Chain of thought & Table \ref{tab:cot_prompt}
\\
Evaluator agents (round 1) & Table \ref{tab:debate_prompt_first_round}
\\
Evaluator agents & Table \ref{tab:debate_prompt}
\\
Adjudicator agents& Table \ref{tab:adjudicator_prompt}
\\
Ambiguity baseline & Table \ref{tab:ambiguity_baseline_prompt}
\\
Ambiguity with debate arguments & Table \ref{tab:ambiguity_w_args_prompt}
\\
\bottomrule
\end{tabular}
\caption{List of prompts used for experiments.}
\label{tab:prompt_lists}
\end{table}
\begin{table*}
\centering
\small
\begin{tabular}{@{}p{14cm}@{}}
\toprule
You are given a document and a summary (summarizing only a part of the document). You will go over the document in the <doc></doc> tags carefully and try to understand it fully. Then you look at the summary in <summary></summary> tags carefully. Your task is to identify whether the summary is factually consistent with the given document. A summary is factually consistent with the document if it can be entailed (either stated or implied) by it.
\newline
\newline
    <doc>
    \newline
    \%s
    \newline
    </doc>
    \newline
    \newline
    <summary>
    \newline
    \%s
    \newline
    </summary>
    \newline
    \newline
    Determine if the summary is factually consistent with the document provided above. You should go over each sentence of the summary one by one and check whether there is an error or not. A summary is non-factual if there is at least one error in it. Provide your evaluation between <label></label> tags with values 1 (consistent) or 0 (inconsistent) and add your explanations in <explanation></explanation> XML tags.
    \\ 
 \bottomrule
\end{tabular}
\caption{Prompt used for zero-shot faithfulness evaluation}
\label{tab:direc_prompt}
\end{table*}
\begin{table*}
\centering
\small
\begin{tabular}{@{}p{14cm}@{}}
\toprule
You are given a document and a summary (summarizing only a part of the document). You will go over the document in the <doc></doc> tags carefully and try to understand it fully. Then you look at the summary sentence in <summary></summary> tags carefully. Your task is to identify whether the summary is factually consistent with the given document. A summary is factually consistent with the document if it can be entailed (either stated or implied) by it.
\newline
\newline
    <doc>
    \newline
    \%s
    \newline
    </doc>
    \newline
    \newline
    <summary>
    \newline
    \%s
    \newline
    </summary>
    \newline
    \newline
    Determine if the sentence is factually consistent with the document provided above. Provide your evaluation between <label></label> tags with values 1 (consistent) or 0 (inconsistent) and add your explanations in <explanation></explanation> XML tags. Before answering, please think about the question within <thinking></thinking> XML tags.
    \\ 
 \bottomrule
\end{tabular}
\caption{Prompt used for chain of thought faithfulness evaluation}
\label{tab:cot_prompt}
\end{table*}
\begin{table*}
\centering
\small
\begin{tabular}{@{}p{14cm}@{}}
\toprule
You are given a document and a summary (summarizing only a part of the document). You will go over the document in the <doc></doc> tags carefully and try to understand it fully. Then you look at the summary sentence in <summary></summary> tags. You have to identify whether the summary is factually consistent with the given document. There are also other evaluator agents assigned the same task as you and you can also see the discussion history in <chat\_history></chat\_history> tags below. You are also given a set of guidelines in <guideline></guidelines> that you can refer to when making your arguments. Go over them carefully and make sure you remember them.
\newline
\newline
<guidelines>
\newline
    1. You should aim for accuracy and not comprehensiveness. If individual facts are correct, the summary is factually consistent regardless of its comprehensiveness.
    \newline
    2. A summary does not imply that its facts are the only ones mentioned in the dialogue.
    \newline
    3. The summary is factually inconsistent if it makes an assumption that is not supported (explicitly or implicitly) by the document.
    \newline
    4. The summary is factually inconsistent if it includes any information (even a minor detail) that is not present in the document or can not be entailed from the document.
    \newline
    5. The summary is factually consistent if it is a paraphrase of the document and it does not change the meaning of what is stated in the document.
    \newline
    6. Details (even crucial) that are present in the document but omitted in the summary do not lead to factual inconsistency.
    \newline
    7. lack of coherence between summary sentences does not necessarily lead to factual inconsistency.
    \newline
    8. The summary should not hallucinate new entities such as new people or locations not mentioned in the document otherwise it is factually inconsistent.
    \newline
    9. The summary does not have to provide the context or focus only on the main points of the document, it can only focus on a minor concept.
    \newline
    10. The summary is factually consistent even if it omits crucial details from document.
    \newline
    11. The addition of details that are not mentioned in the document or can not be entailed from it, makes the summary factually inconsistent.
    \newline
    12. Every word or phrase of the summary (or its paraphrase) should be present in the document otherwise the summary is factually inconsistent.
    \newline
    13. If even a single part of the summary is factually inconsistent, then the whole summary is factually inconsistent.
    \newline
    </guidelines>
\newline
\newline
    <doc>
    \newline
    \%s
    \newline
    </doc>
    \newline
    \newline
    <summary>
    \newline
    \%s
    \newline
    </summary>
    \newline
    \newline
    <chat\_history>
    \newline
    You (Agent 1): The summary is faithful.
    \newline
    Agent 2: The summary is unfaithful.
    \newline
    Agent 3: The summary is faithful.
    \newline
    Agent 4: The summary is unfaithful.
    \newline
    </chat\_history>
    \newline
    \newline
    Determine if the summary is factually consistent with the document provided above. Provide your evaluation between <label></label> tags with values 1 (consistent) or 0 (inconsistent) and add your explanations in <explanation></explanation> XML tags. Before answering, please think about the question within <thinking></thinking> XML tags.
    \\ 
 \bottomrule
\end{tabular}
\caption{Prompt used for evaluator agents for the first round of debate for faithfulness evaluation.}
\label{tab:debate_prompt_first_round}
\end{table*}
\begin{table*}
\centering
\small
\begin{tabular}{@{}p{14cm}@{}}
\toprule
You are given a document and a summary (summarizing only a part of the document). You will go over the document in the <doc></doc> tags carefully and try to understand it fully. Then you look at the summary sentence in <summary></summary> tags. You have to identify whether the summary is factually consistent with the given document. There are also other evaluator agents assigned the same task as you and you can also see the discussion history in <chat\_history></chat\_history> tags below. You are also given a set of guidelines in <guideline></guidelines> that you can refer to when making your arguments. Go over them carefully and make sure you remember them.
\newline
\newline
<guidelines>
\newline
    1. You should aim for accuracy and not comprehensiveness. If individual facts are correct, the summary is factually consistent regardless of its comprehensiveness.
    \newline
    2. A summary does not imply that its facts are the only ones mentioned in the dialogue.
    \newline
    3. The summary is factually inconsistent if it makes an assumption that is not supported (explicitly or implicitly) by the document.
    \newline
    4. The summary is factually inconsistent if it includes any information (even a minor detail) that is not present in the document or can not be entailed from the document.
    \newline
    5. The summary is factually consistent if it is a paraphrase of the document and it does not change the meaning of what is stated in the document.
    \newline
    6. Details (even crucial) that are present in the document but omitted in the summary do not lead to factual inconsistency.
    \newline
    7. lack of coherence between summary sentences does not necessarily lead to factual inconsistency.
    \newline
    8. The summary should not hallucinate new entities such as new people or locations not mentioned in the document otherwise it is factually inconsistent.
    \newline
    9. The summary does not have to provide the context or focus only on the main points of the document, it can only focus on a minor concept.
    \newline
    10. The summary is factually consistent even if it omits crucial details from document.
    \newline
    11. The addition of details that are not mentioned in the document or can not be entailed from it, makes the summary factually inconsistent.
    \newline
    12. Every word or phrase of the summary (or its paraphrase) should be present in the document otherwise the summary is factually inconsistent.
    \newline
    13. If even a single part of the summary is factually inconsistent, then the whole summary is factually inconsistent.
    \newline
    </guidelines>
\newline
\newline
    <doc>
    \newline
    \%s
    \newline
    </doc>
    \newline
    \newline
    <summary>
    \newline
    \%s
    \newline
    </summary>
    \newline
    \newline
    <chat\_history>
    \newline
    \%s
    \newline
    </chat\_history>
    \newline
    \newline
    Determine if the summary is factually consistent with the document provided above. Provide your evaluation between <label></label> tags with values 1 (consistent) or 0 (inconsistent) and add your explanations in <explanation></explanation> XML tags. Before answering, please think about the question within <thinking></thinking> XML tags.
    \\ 
 \bottomrule
\end{tabular}
\caption{Prompt used for evaluator agents during debate for faithfulness evaluation}
\label{tab:debate_prompt}
\end{table*}
\begin{table*}
\centering
\small
\begin{tabular}{@{}p{14cm}@{}}
\toprule
You are given a document, a summary (summarizing only a part of the document) and multiple judgments from evaluator agents. You will go over the document in the <doc></doc> tags and the summary sentence in <summary></summary> tags carefully. A summary is factually consistent if it can be entailed from the document. You go over the discussion between the agents and their arguments shown in between <chat\_history></chat\_history> tags. Your task is to make the final call on whether the summary is factually consistent with the given document based on the evaluator agents responses. You are also given a set of guidelines in <guideline></guidelines> which the agents have referred to, to make their arguments. Go over the guideline carefully and try to remember them.
\newline
\newline
<guidelines>
\newline
    1. You should aim for accuracy and not comprehensiveness. If individual facts are correct, the summary is factually consistent regardless of its comprehensiveness.
    \newline
    2. A summary does not imply that its facts are the only ones mentioned in the dialogue.
    \newline
    3. The summary is factually inconsistent if it makes an assumption that is not supported (explicitly or implicitly) by the document.
    \newline
    4. The summary is factually inconsistent if it includes any information (even a minor detail) that is not present in the document or can not be entailed from the document.
    \newline
    5. The summary is factually consistent if it is a paraphrase of the document and it does not change the meaning of what is stated in the document.
    \newline
    6. Details (even crucial) that are present in the document but omitted in the summary do not lead to factual inconsistency.
    \newline
    7. lack of coherence between summary sentences does not necessarily lead to factual inconsistency.
    \newline
    8. The summary should not hallucinate new entities such as new people or locations not mentioned in the document otherwise it is factually inconsistent.
    \newline
    9. The summary does not have to provide the context or focus only on the main points of the document, it can only focus on a minor concept.
    \newline
    10. The summary is factually consistent even if it omits crucial details from document.
    \newline
    11. The addition of details that are not mentioned in the document or can not be entailed from it, makes the summary factually inconsistent.
    \newline
    12. Every word or phrase of the summary (or its paraphrase) should be present in the document otherwise the summary is factually inconsistent.
    \newline
    13. If even a single part of the summary is factually inconsistent, then the whole summary is factually inconsistent.
    \newline
    </guidelines>
\newline
\newline
    <doc>
    \newline
    \%s
    \newline
    </doc>
    \newline
    \newline
    <summary>
    \newline
    \%s
    \newline
    </summary>
    \newline
    \newline
    <chat\_history>
    \newline
    \%s
    \newline
    </chat\_history>
    \newline
    \newline
    Go over the agents responses, summarize them by saying who agrees/disagrees. Then looking at the agents responses, how well they are associated with the guidelines and finally your own judgement of the summary using the provided guidelines, determine if the summary is factually consistent with the document. Provide your evaluation between <label></label> keys with values 1 (consistent) or 0 (inconsistent) and add your explanations in <explanation></explanation> XML tags.
    \\ 
 \bottomrule
\end{tabular}
\caption{Prompt used for the adjudicator agents.}
\label{tab:adjudicator_prompt}
\end{table*}
\begin{table*}
\centering
\small
\begin{tabular}{@{}p{14cm}@{}}
\toprule
You are given a document and a summary. You will go over the document in the <doc></doc> tags carefully and try to understand it fully. Then you look at the summary in <summary></summary> tags carefully. Your task is to identify whether the summary contains an ambiguity according to the provided ambiguity taxonomy in <taxonomy></taxonomy> tags. A summary is ambiguous if it can have multiple correct interpretations.
\newline
\newline
 <doc>
    \newline
    \%s
    \newline
    </doc>
    \newline
    \newline
    <summary>
    \newline
    \%s
    \newline
    </summary>
    \newline
    \newline
<taxonomy>
\newline
    1. Deduction: The summarizer has made a logical deduction (well or poorly), utilizing premises from the source document to draw a conclusion that cannot be directly traced to the source document.
    \newline
    2. Common-sense inference: The summarizer appears to have made an inference based on common sense notions.
    \newline
    3. Value-based inference: The summarizer appears to have made an inference based on assumed values.
    \newline
    4. Other implicit reasoning phenomenon: Some other kind of implicit reasoning took place that affects the summary's evaluability.
    \newline
    5. Hypernymy/Generalization: A more general meaning is used in the summary than is observed in the source document (for the same topic).
    \newline
    6. Hyponymy/Specialization: A more specific meaning is used in the summary than is observed in the source document (for the same topic).
    \newline
    7. Synonymy/Paraphrasing: Meaning from the source document is paraphrased in such a way that interpretation is challenged. The meaning has not technically changed, but the way the meaning is built changed.
    \newline
    8. Structural ambiguity: A phrase or sentence in the summary has multiple valid parses (multiple valid syntactic structures), and it is not obvious which parse is intended.
    \newline
    9. Lexical ambiguity: A word in the summary has multiple valid interpretations, and it is not obvious which meaning is intended.
    \newline
    10. Other ambiguity phenomenon: There is another type of linguistic ambiguity in the summary that is likely to cause difficulty in interpretation. Other types of ambiguity include scope ambiguity and pronoun reference ambiguity.
    \newline
    11. Vagueness: The meaning of part of the summary is underspecified, resulting in many realities being compatible with the claim made. For this use case, it would be so many realities that there is confusion about what claim is actually being made and whether the claim can be evaluated reliably.
    \newline
    12. Other meaning phenomenon: There is something else about the literal meaning of the summary that may have made it challenging to assess its factuality.
    \newline
    13. Decontextualization: The summary puts forth or describes something outside of the context in which its meaning was meant to be interpreted. It takes on new meaning or loses its meaning outside of that context.
    \newline
    14. Conflation: The summary joins or synthesizes pieces of information that were independently relevant in the source document. (It may have done this to good effect or to bad effect.)
    \newline
    15. Other context phenomenon: Some other challenge related to the relationship between the summary's meaning and the context(s) in the source document.
    \newline
    </taxonomy>
\newline
\newline
Go over the agents responses, summarize them by saying who agrees/disagrees. Then looking at the agents responses, how well they are associated with the guidelines and finally your own judgement of the summary using the provided guidelines, determine if the summary is factually consistent with the document. Provide your evaluation between <label></label> keys with values 1 (consistent) or 0 (inconsistent) and add your explanations in <explanation></explanation> XML tags.
    \\ 
 \bottomrule
\end{tabular}
\caption{Prompt used for ambiguity detection baseline.}
\label{tab:ambiguity_baseline_prompt}
\end{table*}
\begin{table*}
\centering
\scriptsize
\begin{tabular}{@{}p{14cm}@{}}
\toprule
You are given a document and a summary. You will go over the document in the <doc></doc> tags carefully and try to understand it fully. Then you look at the summary in <summary></summary> tags carefully. Evaluator agents have had rounds of discussion to identify whether the summary is factual or not and you can see their arguments in <arguments></arguments> tags. Different agents might have contrasting reasonings on whether the summary is factual or not and they might be correct in their judgement even though they have opposing views. Your task is to go over the arguments and identify whether the summary contains an ambiguity using the provided ambiguity taxonomy in <taxonomy></taxonomy> tags that can cause opposing views of the factuality. An ambiguity is present when the summary can be correctly classified as both factual and non-factual at the same time. Please note that the arguments might not be correct as the agents might have misused the provided guidelines in <guidelines></guidelines> tags so first make sure the agents’ arguments indeed follow the guidelines and then only consider the ones that are sound in your ambiguity evaluation.
\newline
\newline
 <doc>
    \newline
    \%s
    \newline
    </doc>
    \newline
    \newline
    <summary>
    \newline
    \%s
    \newline
    </summary>
    \newline
    \newline
    <arguments>
    \newline
    \%s
    \newline
    </arguments>
    \newline
    \newline
<guidelines>
\newline
    1. You should aim for accuracy and not comprehensiveness. If individual facts are correct, the summary is factually consistent regardless of its comprehensiveness.
    \newline
    2. A summary does not imply that its facts are the only ones mentioned in the dialogue.
    \newline
    3. The summary is factually inconsistent if it makes an assumption that is not supported (explicitly or implicitly) by the document.
    \newline
    4. The summary is factually inconsistent if it includes any information (even a minor detail) that is not present in the document or can not be entailed from the document.
    \newline
    5. The summary is factually consistent if it is a paraphrase of the document and it does not change the meaning of what is stated in the document.
    \newline
    6. Details (even crucial) that are present in the document but omitted in the summary do not lead to factual inconsistency.
    \newline
    7. lack of coherence between summary sentences does not necessarily lead to factual inconsistency.
    \newline
    8. The summary should not hallucinate new entities such as new people or locations not mentioned in the document otherwise it is factually inconsistent.
    \newline
    9. The summary does not have to provide the context or focus only on the main points of the document, it can only focus on a minor concept.
    \newline
    10. The summary is factually consistent even if it omits crucial details from document.
    \newline
    11. The addition of details that are not mentioned in the document or can not be entailed from it, makes the summary factually inconsistent.
    \newline
    12. Every word or phrase of the summary (or its paraphrase) should be present in the document otherwise the summary is factually inconsistent.
    \newline
    13. If even a single part of the summary is factually inconsistent, then the whole summary is factually inconsistent.
    \newline
    </guidelines>
\newline
\newline
<taxonomy>
\newline
    1. Deduction: The summarizer has made a logical deduction (well or poorly), utilizing premises from the source document to draw a conclusion that cannot be directly traced to the source document.
    \newline
    2. Common-sense inference: The summarizer appears to have made an inference based on common sense notions.
    \newline
    3. Value-based inference: The summarizer appears to have made an inference based on assumed values.
    \newline
    4. Other implicit reasoning phenomenon: Some other kind of implicit reasoning took place that affects the summary's evaluability.
    \newline
    5. Hypernymy/Generalization: A more general meaning is used in the summary than is observed in the source document (for the same topic).
    \newline
    6. Hyponymy/Specialization: A more specific meaning is used in the summary than is observed in the source document (for the same topic).
    \newline
    7. Synonymy/Paraphrasing: Meaning from the source document is paraphrased in such a way that interpretation is challenged. The meaning has not technically changed, but the way the meaning is built changed.
    \newline
    8. Structural ambiguity: A phrase or sentence in the summary has multiple valid parses (multiple valid syntactic structures), and it is not obvious which parse is intended.
    \newline
    9. Lexical ambiguity: A word in the summary has multiple valid interpretations, and it is not obvious which meaning is intended.
    \newline
    10. Other ambiguity phenomenon: There is another type of linguistic ambiguity in the summary that is likely to cause difficulty in interpretation. Other types of ambiguity include scope ambiguity and pronoun reference ambiguity.
    \newline
    11. Vagueness: The meaning of part of the summary is underspecified, resulting in many realities being compatible with the claim made. For this use case, it would be so many realities that there is confusion about what claim is actually being made and whether the claim can be evaluated reliably.
    \newline
    12. Other meaning phenomenon: There is something else about the literal meaning of the summary that may have made it challenging to assess its factuality.
    \newline
    13. Decontextualization: The summary puts forth or describes something outside of the context in which its meaning was meant to be interpreted. It takes on new meaning or loses its meaning outside of that context.
    \newline
    14. Conflation: The summary joins or synthesizes pieces of information that were independently relevant in the source document. (It may have done this to good effect or to bad effect.)
    \newline
    15. Other context phenomenon: Some other challenge related to the relationship between the summary's meaning and the context(s) in the source document.
    \newline
    </taxonomy>
\newline
\newline
Go over the agents responses, summarize them by saying who agrees/disagrees. Then looking at the agents responses, how well they are associated with the guidelines and finally your own judgement of the summary using the provided guidelines, determine if the summary is factually consistent with the document. Provide your evaluation between <label></label> keys with values 1 (consistent) or 0 (inconsistent) and add your explanations in <explanation></explanation> XML tags.
    \\ 
 \bottomrule
\end{tabular}
\caption{Prompt used for ambiguity detection with debate arguments.}
\label{tab:ambiguity_w_args_prompt}
\end{table*}



\section{Additional Results}\label{app:result}
\subsection{\method shows similar trends on sentence-level summaries.}\label{app:full_results}
Table \ref{tab:full} presents a mix of finetuned and LLM-based evaluators along with our debate variants both on sentence-level and summary-level faithfulness evaluation.
\begin{table*}
\scriptsize
\centering
\begin{tabular}{@{}clcccccccccccc@{}}
\toprule
\multirow{4}{*}{LLM}  & \multirow{4}{*}{Model} & \multicolumn{8}{c}{TofuEval}                                                                                                                    & \multicolumn{4}{c}{AggreFact}                                                        \\ \cmidrule(l){3-13} 
                       & & \multicolumn{4}{c}{MeetingBank}                                        & \multicolumn{4}{c}{MediaSum}                                           & \multicolumn{2}{c}{\multirow{2}{*}{CNN}} & \multicolumn{2}{c}{\multirow{2}{*}{XSum}} \\
                       \cmidrule(l){3-6}\cmidrule(l){7-10}
                       & & \multicolumn{2}{c}{Sentence-Level} & \multicolumn{2}{c}{Summary-Level} & \multicolumn{2}{c}{Sentence-Level} & \multicolumn{2}{c}{Summary-Level} & \multicolumn{2}{c}{}                     & \multicolumn{2}{c}{}                      \\ \cmidrule(l){3-4}\cmidrule(l){5-6}\cmidrule(l){7-8}\cmidrule(l){9-10}\cmidrule(l){11-12}\cmidrule(l){13-14}
                       & & BAcc           & K-alpha           & BAcc           & K-alpha          & BAcc           & K-alpha           & BAcc           & K-alpha          & \multicolumn{1}{l}{BAcc}    & K-alpha    & BAcc               & K-alpha

\\ \cmidrule(r){1-14}

\multirow{8}{*}{\rotatebox{90}{finetuned}} &  SummaC-CV & 62.80 & - & - & - &63.70& -& - & -& 65.20 & - &54.50 & -
\\
& T5-NLI-Mixed & 55.30 & - & - & - &59.10 & -& - & -& 54.60 & - &52.30 & -
\\
& FT5-ANLI-L & 60.10 & - & - & - &57.40 & -& - & -& 51.20 & - &60.00 & -
\\
& DAE & 69.50 & - & - & - &65.10 & -& - & -& 50.80 & - &59.10 & -
\\
& QAFactEval & 65.70 & - & - & - &61.30 & -& - & -& 54.30 & - &62.10 & -
\\
& SummaC-ZS & 71.00 & - & - & - &69.50 & -& - & -& 51.10 & - &61.50 & -
\\
& AlignScore & 72.60 & - & - & - &69.20 & -& - & -& 52.40 & - &71.40 & -
\\ 
&MiniCheck (Flan-t5) & 77.30 & 0.51 &68.07 &0.30 &73.58 &0.44 & 69.52 &0.36 & 69.95 &0.33 &74.26 &0.48
\\
\cmidrule(r){1-14}
\multirow{6}{*}{\rotatebox{90}{Llama3}}            
&Zero-shot single LLM& 75.57 & 0.52 & 68.15 & 0.38 & 66.09 & 0.38 & 56.23 &0.00 & 60.18 &	0.28	&68.13&	0.35
 \\
&Zero-shot Chain of Thought&75.63 & 0.53 & 68.45 & 0.39 & 65.91 & 0.37 &58.77 & 0.09 & 63.34 & 0.35 & 68.17 &0.35
\\
&Self-consistency &74.71 & 0.52 & 69.05 & 0.40 & 67.14 & 0.41 & 61.07 &0.15
& 62.56 &	0.34&	68.87&	0.37
\\
\cmidrule(r){2-14}
&\method & 79.67 & 0.53 & 75.08 & 0.50 & 75.17 & 0.51 & 68.06& 0.36
 &	66.88	& 0.34 &	75.10 &	0.50
\\
\cmidrule(r){2-14}
&\method w. sim debates (agents vote) & 79.07 & 0.53 & 78.06 & 0.57 & 76.94 & 0.54 & 70.59 & 0.42& 69.13	& 0.39 &	73.62	& 0.47
 \\
&\method w. sim debates (debates vote) & 	79.13 & 0.54 & 77.42 & 0.56 & 76.27& 0.53 & 69.25 & 0.39 & 69.03	& 0.39	& 74.71 & 	0.49
 \\
\bottomrule
\end{tabular}
\caption{Full results on a diversity of fact-checkers both on sentence-level and summary-level summaries. The finetuned results are directly presented from \citet{tang2024minicheck} along with their best performing MiniCheck variant.}
\label{tab:full}
\end{table*}

\subsection{Initialization distribution effect on \method }\label{app:init}
We use a uniform distribution to assign initial stances to the evaluator agents. 
The reason behind doing so is that
since we the instances are random (without knowing what the correct label is) and to have a fair debate without one stance being stronger than the other (by having more agents start with that stance), we decided to have the same number of agents pro each stance.
\begin{table*}
\centering
\begin{tabular}{@{}lcccc@{}}
\toprule
Model                                    & BAcc  & K-alpha & FPR (\%) & FNR (\%) \\ \midrule
\method wo. random initialization   & 63.13 & 0.20    & 1.33     & 72.41    \\
\method & 68.06 & 0.36    & 17.33    & 46.55    \\
\method w. simultaneous debates (4 agents, 2+, 2-)          & 70.59 & 0.42    & 14.00    & 44.83    \\
\method w. simultaneous debates (5 agents, 2+, 3-)          & 69.80 & 0.40    & 23.33    & 37.07    \\
\method w. simultaneous debates (5 agents, 3+, 2-)          & 62.22 & 0.19    & 4.00     & 71.57    \\ \bottomrule
\end{tabular}
\caption{The effect of stance distribution on performance on MediaSum dataset.}
\label{tab:number_of_agents}
\end{table*}
We performed an analysis on how changing the balance of evaluators can affect the performance of \method in Table \ref{tab:number_of_agents}. 
As can be seen in Table \ref{tab:number_of_agents}, the uniform distribution performs the best. Having more agents with initial belief that the summary is unfaithful will result in the lowest FNR but highest FPR as this setup tends to identify more errors. On the other hand, more positive agents fail to identify errors (similar to the setup without initialization).

\subsection{\method can improve FNR.}\label{app:results:fpr-fnr}
We compare the FPR and FNR of different approaches in Table \ref{tab:llama-main-fpr-fnr} and show the decrease in FNR using the debate approach.
\begin{table*}
\centering
\small
\begin{tabular}{@{}lcccccccc@{}}
\toprule
\multirow{3}{*}{\textbf{Model}} & \multicolumn{4}{c}{\textbf{TofuEval}}                                            & \multicolumn{4}{c}{\textbf{AggreFact}}                                 \\ \cmidrule(l){2-5} \cmidrule(l){6-9} 
                                & \multicolumn{2}{c}{\textbf{MeetingBank}} & \multicolumn{2}{c}{\textbf{MediaSum}} & \multicolumn{2}{c}{\textbf{CNN}}   & \multicolumn{2}{c}{\textbf{XSum}} \\
                                \cmidrule(r){2-3}\cmidrule(r){4-5} \cmidrule(r){6-7} \cmidrule(r){8-9}
                                & \multicolumn{1}{l}{FPR}    & FNR    & FPR             & FNR            & \multicolumn{1}{l}{FPR} & FNR & FPR           & FNR        

\\ \cmidrule(r){1-9}
Zero-shot single LLM& 6.55	& 57.14	& 0.01 &	86.20	& 0.80	& 78.95	& 16.49&	47.25
 \\
Zero-shot Chain of Thought& 5.95	& 57.14	& 4.00 &	78.45	& 1.40	&71.93&	18.24	& 45.42
\\
Self-consistency &4.76	&57.14	&2.00&	75.56	&1.20&	73.68&	16.84&	45.42
\\
\cmidrule(r){1-9}
\method wo initialization & 3.58	& 58.16&	1.33&	72.41&	1.00	&78.95	&9.82&	49.82
 \\
\method & 25.00	& 26.50	&16.00	& 50.86	& 4.59 &	56.14	& 30.88	&25.27
 \\
\cmidrule(r){1-9}
\method w. simultaneous debates (agents vote) & 12.43 &	29.59 &	16.67&	44.83	&5.59&	56.14	&24.56	&28.20
 \\
\method w. simultaneous debates (debates vote) & 12.50 &	32.65	& 14.00	& 44.83	& 5.79	&56.14	&24.21	&26.37
 \\
\bottomrule
\end{tabular}
\caption{The FPR and FNR of different evaluators.}
\label{tab:llama-main-fpr-fnr}
\end{table*}
The debate approach can help with identifying more errors as shown by lower FNR in Table \ref{tab:llama-main-fpr-fnr}.  
However, since the debate approach is more sensitive to the errors, the FPR is also increased. 

There are a few hypothesis to describe this phenomena. First, the initialization would increase the evaluator sensitivity to the potential errors which could lead to labeling cases as erroneous for some superficial reasons as ``lack of context'' or ``omission of details''. One way to resolve this issue is to further curate the guidelines that are given to the evaluators during the debate.
Another cause for this increase is the ambiguities in the summaries that would lead to disagreement between human judgments and model judgments. As discussed earlier, ambiguity can be a major source of disagreement on faithfulness evaluation and has to be dealt with before faithfulness evaluation.  
We later show (\ref{app:results:fpr-fnr-filtered}) that filtering ambiguous cases would lower the gap in terms of FPR between the debate approach and other baseline settings.

\subsection{\method is orthogonal to the underlying LLM.}\label{app:results:gpt4}
The comparison of \method with other baselines using GPT-4o-mini as the underlying LLM is shown in Table \ref{tab:gpt-main}.
\begin{table*}
\centering
\resizebox{\textwidth}{!}{
\begin{tabular}{@{}lcccccccc@{}}
\toprule
\multirow{3}{*}{\textbf{Model}} & \multicolumn{4}{c}{\textbf{TofuEval}}                                            & \multicolumn{4}{c}{\textbf{AggreFact}}                                 \\ \cmidrule(l){2-5} \cmidrule(l){6-9} 
                                & \multicolumn{2}{c}{\textbf{MeetingBank}} & \multicolumn{2}{c}{\textbf{MediaSum}} & \multicolumn{2}{c}{\textbf{CNN}}   & \multicolumn{2}{c}{\textbf{XSum}} \\
                                \cmidrule(r){2-3}\cmidrule(r){4-5} \cmidrule(r){6-7} \cmidrule(r){8-9}
                                & \multicolumn{1}{l}{BAcc}    & K-alpha    & BAcc             & K-alpha            & \multicolumn{1}{l}{BAcc} & K-alpha & BAcc           & K-alpha        

\\ \cmidrule(r){1-9}
Zero-shot single LLM& 69.30	& 0.41 &	62.07&	0.16&	58.17&	0.22&	72.00&	0.44
\\
Self-consistency (n=40) &68.62	&0.39	&65.51	&0.26	&56.42&	0.17&	74.63	& 0.49
 \\
\cmidrule(r){1-9}
\method & 74.40	&0.46	&68.05	&0.36	&70.79	&0.13	&72.86	&0.45
\\
\cmidrule(r){1-9}
\method w. simultaneous debates (agents vote) & 76.96	&0.54&	72.51	&0.46	&70.63	&0.33&	74.35&	0.49
\\
\method w. simultaneous debates (debates vote) & 77.76	& 0.56 &	72.94 &	0.47	&71.30&	0.33	&74.53&	0.49
\\
\bottomrule
\end{tabular}
}
\caption{Main table comparing the debate setup with baselines using GPT-4o-mini as the main LLM.}
\label{tab:gpt-main}
\end{table*}
Though self-consistency on XSum dataset is the highest performing baseline, its performance is not even close to any debate settings for other datasets in Table \ref{tab:gpt-main}.

\subsection{\method works for smaller LLMs as well}\label{app:results:llama-small}
We also show in Table \ref{tab:llama-small-main} that \method can be superior to baselines even when a smaller size LLM is used.
Even though the debate setup for a smaller-size LLM does not reach the larger LLM performance in Table \ref{tab:llama-main}, but it can beat any other single LLM-based approaches using the larger LLM.
\begin{table*}
\centering
\begin{tabular}{@{}lcccc@{}}
\toprule
\multirow{2}{*}{\textbf{Model}} & \multicolumn{4}{c}{\textbf{MediaSum}} \\ \cmidrule(l){2-5} 
                                & BAcc  & K-alpha & FPR (\%) & FNR (\%)      

\\ \cmidrule(r){1-5}
Zero-shot single LLM& 55.56 &	-	&2.67 &	86.21
 \\
\cmidrule(r){1-5}
\method & 58.92	&0.18	&37.33	&44.83
\\
\cmidrule(r){1-5}
\method w. simultaneous debates (agents vote) & 61.81	&0.21&	10.00&	66.38
 \\
\method w. simultaneous debates (debates vote) & 63.10	& 0.24 &	10.00 &	63.79
 \\
\bottomrule
\end{tabular}
\caption{Main table comparing the results on a small size model Llama-3-8b.}
\label{tab:llama-small-main}
\end{table*}


\subsection{Ambiguity filtering can help with balancing FPR and FNR.}\label{app:results:fpr-fnr-filtered}
\begin{figure*}
\large
\begin{minipage}{0.49\textwidth} \centering
\begin{tikzpicture}[scale=0.7]
    \begin{axis}[
       ybar=2*\pgflinewidth,
        bar width=0.8cm,
        ymajorgrids = true,
        grid style=dashed,
        ylabel={BAcc},
        nodes near coords,
        symbolic x coords={Zero-shot,Consistency,Debate},
        xtick = data,
        scaled y ticks = false,
        enlarge x limits=0.25,
        ymin=4,
        ymax=15,
        legend columns=2,
        legend style={at={(0.05,-0.15)},anchor=north west},
    ]

    \addplot[style={black,pattern color=red,pattern = north west lines}]
    coordinates {(Zero-shot,8.20)(Consistency,7.23)(Debate,13.98)};
    
    \addplot[style={black,pattern color=teal,pattern = crosshatch}]
    coordinates {(Zero-shot,5.16)(Consistency,4.96)(Debate,7.14)};

\addlegendentry{Unfiltered}
\addlegendentry{Filtered}

    \end{axis}
\end{tikzpicture}
\end{minipage}\hfill
\begin{minipage}{0.49\textwidth} \centering
\begin{tikzpicture}[scale=0.7]
    \begin{axis}[
       ybar=2*\pgflinewidth,
        bar width=0.8cm,
        ymajorgrids = true,
        grid style=dashed,
        ylabel={K-alpha},
        nodes near coords,
        symbolic x coords={Zero-shot,Consistency,Debate},
        xtick = data,
        scaled y ticks = false,
        enlarge x limits=0.25,
        ymin=15,
        ymax=47,
        legend columns=2,
        legend style={at={(0.05,-0.15)},anchor=north west},
    ]

    \addplot[style={black,pattern color=red,pattern = north west lines}]
    coordinates {(Zero-shot,40.67)(Consistency,43.33)(Debate,26.67)};
    
    \addplot[style={black,pattern color=teal,pattern = crosshatch}]
    coordinates {(Zero-shot,29.66)(Consistency,33.05)(Debate,17.80)};

\addlegendentry{Unfiltered}
\addlegendentry{Filtered}
\end{axis}
\end{tikzpicture}
\end{minipage}\hfill

\caption{FPR and FNR results pre and post filtering the ambiguous cases on annotated (with ambiguity annotation) MeetingBank dataset.}
  \label{plot:ambiguity-filtering-fpr-fnr}
\end{figure*}
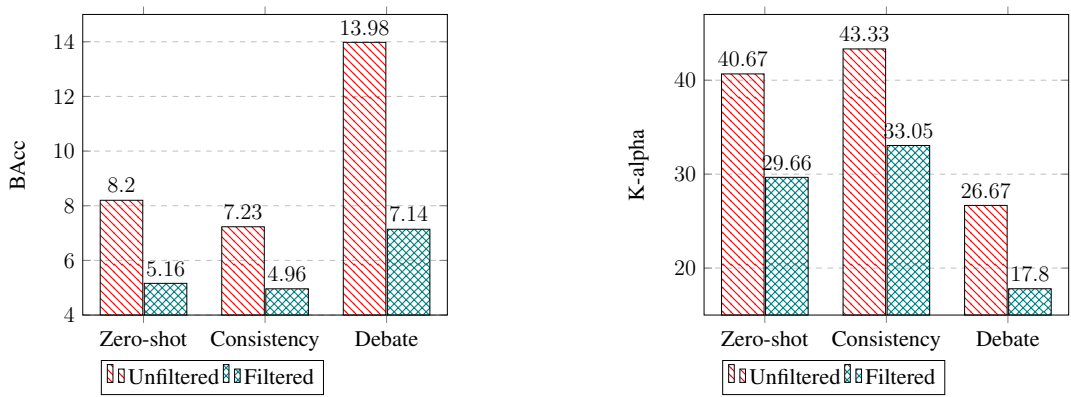

We previously observed that with \method we have lower FNR rate however, the FPR is also increased. 
Once the ambiguous cases are filtered, we can see the decrease in FPR as well. This further suggests that our assumption on how the ambiguous cases can lead to higher FPR is true.
Figure \ref{plot:ambiguity-filtering-fpr-fnr} shows the decline in both FPR and FNR. The FPR gap between the debate approach and different setups is lower once ambiguous cases are filtered.
\end{document}